\documentclass[runningheads]{llncs}
\usepackage{graphicx}
\usepackage{amsmath,bm}
\usepackage{tikz}
\usepackage{comment}
\usepackage{amsmath,amssymb} 
\usepackage{color}
\usepackage[pagebackref,breaklinks,colorlinks]{hyperref}
\usepackage[accsupp]{axessibility}  
\usepackage{booktabs}

\usepackage{amsmath}
\usepackage{amssymb}
\usepackage{booktabs}
\usepackage{enumitem}
\usepackage{multirow}
\usepackage{makecell}
\usepackage{graphicx}
\usepackage{diagbox}
\usepackage{tabu}
\usepackage{rotating}
\usepackage{array,multirow}
\usepackage{float}
\usepackage{enumitem}
\usepackage{wrapfig}
\usepackage{pifont}
\usepackage{algorithm}
\usepackage{amsmath}
\usepackage{amssymb}
\usepackage{booktabs}
\usepackage{algpseudocode} 
\usepackage{color}
\usepackage{enumitem}
\usepackage{eucal}

\def\eg{\emph{e.g.}} 
\def\ie{\emph{i.e.}}

\def\etal{\emph{et al}}

\begin{document}
\pagestyle{headings}
\mainmatter
\def\ECCVSubNumber{6326}  

\title{Identifying Hard Noise in Long-Tailed Sample Distribution} 


\titlerunning{Identifying Hard Noise in Long-Tailed Sample
Distribution}
%
\author{
    Xuanyu Yi\inst{1},
    Kaihua Tang\inst{1}\thanks{Corresponding author.}, 
    Xian-Sheng Hua\inst{2},
    Joo-Hwee Lim\inst{3}, 
    Hanwang Zhang\inst{1}
}
\authorrunning{X. Yi et al.}
%
\institute{
    Nanyang Technological University, Singapore \and
    Damo Academy, Alibaba Group, Hangzhou, China \and
     Institute for Infocomm Research, Singapore \\
    \email{xuanyu001@e.ntu.edu.sg}; \email{kaihua.tang@ntu.edu.sg}; \email{ xshua@outlook.com} \\
    \email{ joohwee@i2r.a-star.edu.sg}; \email{hanwangzhang@ntu.edu.sg}
}

\maketitle

\begin{abstract}
Conventional de-noising methods rely on the assumption that all samples are independent and identically distributed, so the resultant classifier, though disturbed by noise, can still easily identify the noises as the outliers of training distribution. However, the assumption is unrealistic in large-scale data that is inevitably long-tailed. Such imbalanced training data makes a classifier less discriminative for the tail classes, whose previously ``easy'' noises are now turned into ``hard'' ones---they are almost as outliers as the clean tail samples. We introduce this new challenge as Noisy Long-Tailed Classification (NLT). Not surprisingly, we find that most de-noising methods fail to identify the hard noises, resulting in significant performance drop on the three proposed NLT benchmarks: ImageNet-NLT, Animal10-NLT, and Food101-NLT. To this end, we design an iterative noisy learning framework called Hard-to-Easy (H2E). Our bootstrapping philosophy is to first learn a classifier as noise identifier \emph{invariant} to the class and context distributional changes, reducing ``hard'' noises to ``easy'' ones,  whose removal further improves the invariance. Experimental results show that our H2E outperforms state-of-the-art de-noising methods and their ablations on long-tailed settings while maintaining a stable performance on the conventional balanced settings. Datasets and codes are available at \url{https://github.com/yxymessi/H2E-Framework}.

\end{abstract}

\section{Introduction}
\label{sec:1}


Any visual model should learn to co-exist with noise because any real-world dataset is imperfect~\cite{northcutt2021confident}. During data collection, noise such as sensory failure (\eg, low-quality or corrupted images) and human error (\eg, mislabeling or ambiguous annotation) may hurt model training. In general, noise can be understood as a small population of training samples whose image contents differ from the ground-truth classes~\cite{frenay2013classification}. Therefore, if the data is independent and identically distributed (IID) regardless of class~\cite{rolnick2017deep,fawzi2016robustness,sastry2017robust}, noise samples can be identified as the outliers of the classifier confidence~\cite{han2018co,chen2015webly,chen2019understanding}. Specifically, we first learn the classifier on noisy data, then identify the noises as outliers, and finally remove them for a cleaner data that improves the classifier---a virtuous cycle~\cite{li2020dividemix,yu2019does}. In particular, we term the noise that can be identified as outliers as ``easy'' noise.

\begin{figure}[t]
   \begin{minipage}[b]{1.0\linewidth}
   \centerline{\includegraphics[scale=0.27 
   ]{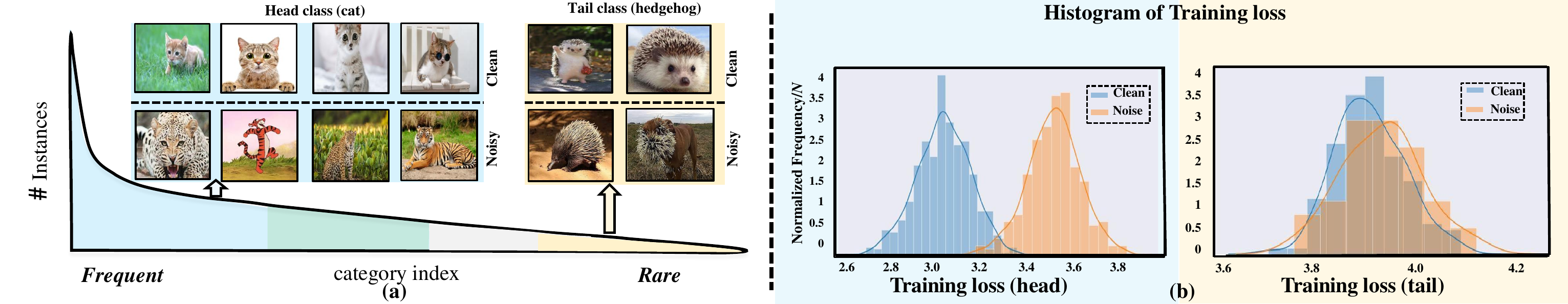}}
   \end{minipage}
   \caption{(a) Large-scale datasets are both long-tailed and noisy. For instance, a head category ``cat'' may contain noisy samples such as ``leopard'' and ``cartoon tiger'' while noise like ``porcupine'' and ``spiny horse'' in tail category ``hedgehog''  (b) The identification of noise based on classifier confidence (or training loss) is no longer applicable in tail classes for most de-noise algorithms}
   \label{fig:1}
\end{figure}

On a dataset with the balanced number of diverse training samples per class---the conventional settings as in most de-noise literature~\cite{li2020dividemix,mirzasoleiman2020coresets,chen2021boosting}---the IID assumption is easy to be satisfied. The key reason is that such dataset can guarantee a robust classifier that only focuses on the context-invariant class feature (or causal feature)~\cite{arjovsky2019invariant,rosenfeld2020risks,qi2022class,wang2022equivariance}. Therefore, after learning to exclude all the varying contexts (non-causal features), the class features of clean and noisy samples are indeed different. For example, even if the noise is as tricky as a ``leopard'' sample mislabeled as ``cat'' that is visually similar to ``leopard'', after removing the context, ``cat'' feature is still different from ``leopard'' feature, who is an ``easy'' outlier of ``cat''. So, the premise of the IID assumption is the disentanglement of the class and context features.

However, should a dataset be at scale, long-tailed distribution will be inevitable~\cite{zhang2021deep,lee2021learning}, and thus disentangling class and context becomes challenging. The reasons are due to the following two folds that make the classifier dependent on \textbf{class prior} and \textbf{context distribution}. \textit{First}, as head has more samples than tail, the classifier will be biased to head~\cite{tang2020long}. \textit{Second}, head samples have more diverse contexts than tail, \ie, contexts are not shared by all the classes and some contexts are unique to certain tail classes due to sample scarcity. So, the resultant classifier fails to learn context-invariant class features, but entangling context with class~\cite{wang2021self}. As shown in Fig.~\ref{fig:1}(a), the ``spine'' context is highly correlated to ``hedgehog'' and thus ``spine'' is a confusing context to mis-recognize the noise ``porcupine'' and ``spiny horse'' as ``hedgehog''. Thus the long-tailed distribution will turn ``easy'' noise into ``hard'', especially for the tail classes. Fig.~\ref{fig:1}(b) illustrate such an example: the noises in tail class are almost as outliers as the tail samples. We leave a more detailed analysis in Section~\ref{sec:3}.

In this paper, we present a new challenge for noisy learning at scale, called Noisy Long-Tailed classification (NLT), which unifies the long-tailed distribution with realistic noisy data, completing the pioneering work with only synthetic noise on imbalanced data~\cite{wei2021robust,cao2020heteroskedastic,karthik2021learning}. For rigorous and reproducible evaluations in the community, we introduce three benchmarks: ImageNet-NLT, Animal10-NLT, and Food101-NLT, with various noise and imbalance ratio for comprehensive diagnosis (Section~\ref{sec:5}). Not surprisingly, most of the existing de-noise methods degrade significantly on the benchmarks, especially for those who heavily rely on outlier detection~\cite{jiang2018mentornet,yu2019does,li2020dividemix,wu2020topological}.

\begin{wrapfigure}{r}{3.5cm}
\includegraphics[width=3.5cm]{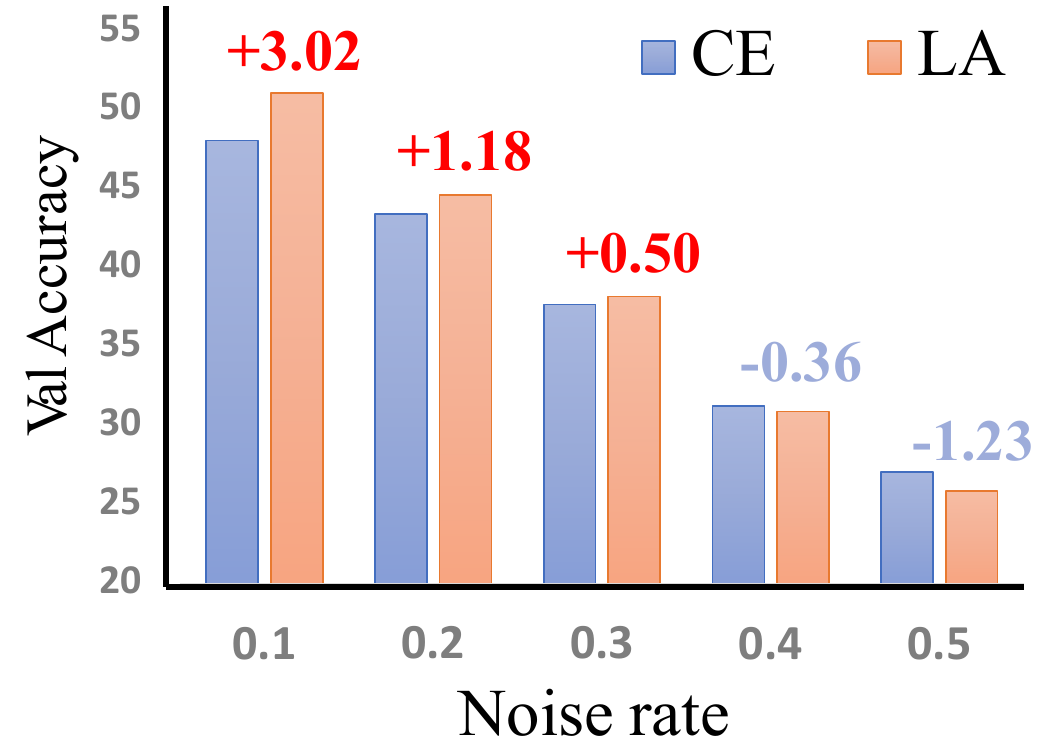}
\caption{\scriptsize{The comparison of CE (cross-entropy)~\cite{shore1981properties} and Logit-Adjustment~\cite{menon2020long} in CIFAR-100 with different noise ratios.}}
\label{fig:2}
\end{wrapfigure}

One may wonder if we could first learn a balanced classifier on the noisy data by using long-tailed classification methods~\cite{zhou2020bbn,tang2020long,kang2019decoupling}, and then apply the conventional outlier detection for noise identification. The answer is ``No'' because those methods can only mitigate the class bias but not the context bias in long-tailed data. Fig.~\ref{fig:2} demonstrates that with the increase of noise ratio, the performance of a SOTA long-tailed method~\cite{menon2020long} decreases significantly.




To this end, we propose an iterative Hard-to-Easy (H2E) framework for NLT. It has two stages: 1) A noise identifier that is invariant to the class and context distributional change caused by long-tailed distribution (Section~\ref{sec:4.1}). Such invariance can reduce ``hard'' noises to ``easy'' ones. Specifically, we sample three data distribution: long-tailed, balanced, and reversed long-tailed, as three context environments, and then apply Invariant Risk Minimization (IRM)~\cite{arjovsky2019invariant} to learn a long-tailed classifier as the noise identifier invariant to these environments. Note that this stage is iterative as ``clearner'' data improves training better backbones. 2) Thanks to the noise identifier, we can eventually learn a robust classifier.


Our contributions are summarized as follows:
\begin{itemize}
\setlength\itemsep{0em}
\item We present the task: Noisy Long-Tailed classification (NLT) with non-synthetic real-world noises. NLT is challenging because it turns ``easy'' noises into ``hard'' ones that cannot be identified by prior work.

\item We propose a strong NLT baseline: Hard-to-Easy (H2E) noisy learning framework. The success of H2E is based on learning a noise identifier invariant to the class and context changes introduced by long-tailed data. 

\item We introduce three NLT benchmarks: ImageNet-NLT, Animal10-NLT, and Food101-NLT. Extensive experimental results on them show the limitations of existing de-noise method and the potential of learning invariance for noisy learning.

\end{itemize}

\section{Related Work}
\label{sec:2}

\noindent\textbf{Long-Tailed Classification}. Most existing long-tailed methods can be categorized into three types: 1) class-wise re-balancing using re-sampling strategies~\cite{zhou2020bbn,kang2019decoupling}, re-weighted losses~\cite{ren2020balanced,jamal2020rethinking,shu2019meta}, and post-hoc adjustments~\cite{menon2020long,tang2020long}, 2) data augmentation~\cite{cubuk2020randaugment,liu2020deep}, and 3) model ensembling~\cite{zhang2021test,wang2020long}. Since the latter two aim to boost the overall performance by directly increasing the model capability, which is generally suitable for all classification tasks, rather than focusing on tackling the imbalance between head and tail, we mainly focus on the class-wise re-balancing methods in this paper. Besides, the performance of conventional long-tailed algorithms may significantly degrade in the noisy environment, as they assume the training samples 
to be annotated correctly, which is impractical in real-world images at scale. 


\noindent\textbf{Noisy Learning.} The previous learning with noise algorithms can be summarized into 1) the noisy sample selection~\cite{jiang2018mentornet,han2018co,li2020dividemix} and 2) the regularization~\cite{liu2020peer,kumar2020robust,amid2019robust}. Since the latter methods are generally applicable for all classification tasks, we mainly investigate the former in this paper, as they are more related to the proposed H2E framework. Most of the noisy sample selection methods filter out noisy samples by adopting the \textit{small-loss trick}, which treats samples with small training losses as correct-annotated. In particular, Co-teaching~\cite{han2018co} trains two networks simultaneously where each network selects small-loss samples in a mini-batch to train the other. Beta Mixture Model (BMM)~\cite{arazo2019unsupervised} separates the clean and noise samples during training based on the loss value of each sample. Similarly, Divide-Mix~\cite{li2020dividemix} fits a Gaussian mixture model on per-sample loss distribution to divide the training samples into clean set and noisy set. However, in the existence of the class imbalance, most of these methods may not work well since the training loss converges easier in major classes than minor classes, resulting in the risk of discarding most samples in the minor classes.



\section{Noisy Long-Tailed Classification}
\label{sec:3}

The classification of an image $x$ as class $c$ can be defined as predicting $p(y=c|x)$ based on a dataset of image and ground-truth label pairs $\{(x,y)\}$~\cite{goodfellow2014generative}, where the noise is caused by the wrong label assignment $\widetilde{y} \to x$ ( $\widetilde{y} \neq y$). By Bayes theorem~\cite{bernardo2009bayesian}, we can decompose the predictive model as $p(y=c|x) =  \frac{p(x|y=c) \cdot p(y=c)}{p(x)}$, where $p(y=c)$ is the class distribution, $p(x)$ is the marginal distribution of images. In the independent and identical distribution (IID) assumption of uniform $p(x)$ and $p(y=c)$, it is relatively easy to obtain an \emph{ideal} noise identifier: the classifier $p(y=c|x)$ \emph{per se}, which will be explained later.
I

Unfortunately, the IID assumption is not practical in general as large-scale dataset is usually imbalanced in not only class distribution, but also context distribution. We assume that any image $x$ is generated by a set of hidden semantics $z=\{z_{1},z_{2},z_{3},...\}$, which includes two disjoint subsets: class-specific attributes $z_c$ (\eg, the cat-like shape in the ``cat'' category) and context-specific environmental attributes $z_e$ (\eg, the fur color). So, we can further decompose the predictive model $p(y=c|x = (z_c,z_e))$ as follows:

\begin{equation}
    \label{eq.1}
     p(y=c|z_c,z_e) = \frac{p(z_c|y=c)}{p(z_c,z_e)} \cdot \overbrace{p(z_e | y=c, z_c)}^{context\;bias}  \cdot \overbrace{p(y=c)}^{class\;bias}.
\end{equation}
From Eq.~\eqref{eq.1}, the noise identifier $p(y=c|z_c,z_e)$ is affected by the variations of 1) class bias $p(y=c)$: the distribution shift caused by class imbalance, and 2) context bias $p(z_e|y=c,z_c)$: spurious correlation~\footnote{ \scriptsize{For a thought example based on Eq.~\eqref{eq.1}, if a class-specific attribute ``body'' and a context-specific attribute ``spine'' have strong co-occurrence under the ``hedgehog'' class, the wrong annotation ``hedgehog'' of a ``porcupine'' image with ``spine'' could be imperceptible for the identifier due to the high spurious correspondence $p(z_e=``spine"|y=``hedgehog",z_c=``body")$.}} between context attributes and class.
Such negative effect motivates us to introduce the concept of ``hard'' and ``easy'' noise, which has not been addressed in the de-noise literature yet.

\noindent\textbf{Noise} is defined as training samples with a mismatch between the ground-truth label $y$ and class-specific (causal) features $z_c$.

\noindent\textbf{``Easy'' Noise} could be easily detected by the ideal identifier $p(y=c|z_c,z_e)$, regardless of the influence by $p(z_e | y=c, z_c) \cdot p(y=c)$. That is to say $z_e$ is independent of $y$, \ie, $p(z_e|y=c,z_c)$ approaching $p(z_e|z_c)$ and $z_e$ can be eliminated by  $p(z_e|z_c) / p(z_e,z_c) = 1 / p(z_c)$. Meanwhile, $p(y=c)$ is uniformly distributed under IID assumption in the conventional de-noise setting~\cite{jiang2018mentornet,han2018co,li2020dividemix}. Since the above $1 / p(z_c)$ and $p(y=c)$ could be both considered as constant, noise can be easily identified because $p(y=c|z_c,z_e)$ is directly calculated through the observation of $p(z_c|y=c)$.

\noindent\textbf{``Hard'' Noise} is elusive as $p(y=c|z_c,z_e)$ is affected by the negative impact of $p(z_e | y=c, z_c) \cdot p(y=c)$, leading to erroneous abnormal identification. 

%
 

The proposed \textbf{N}oisy \textbf{L}ong-\textbf{T}ailed classification aims to learn from the training data that possesses two joint phenomena: 1) the class distribution $p(y=c)$ is long-tailed; 2) part of the training samples (noise) are wrongly annotated. Some previous ``easy'' noises are thus turned into ``hard'' ones, resulting in that most of the conventional noise removal algorithms~\cite{northcutt2021confident,arazo2019unsupervised,chen2015webly} are no longer reliable in NLT since the outlier samples can be either caused by the noisy labels with lower $p(z_c|y=c)$ or rare contexts and classes with lower $p(z_e | y=c, z_c) \cdot p(y=c)$. Therefore, we propose the following Hard-to-Easy framework, aiming to learn a fair noise identifier invariant to the change of $p(z_e | y=c, z_c) \cdot p(y=c)$, so the ``hard'' noises can thus be converted into ``easy'' ones.

\section{Hard-to-Easy (H2E) Framework}
\label{sec:4}

As shown in Algorithm~\ref{algo:1}, our H2E framework is composed of two stages with an initial warm-up stage, where \textbf{Stage 1} (Section~\ref{sec:4.1}) obtains a fairer identifier by turning ``hard'' noise into ``easy'' through invariant muti-environment learning, thus obtaining a ``cleaner'' representation by removing the identified ``easy'' noise. An iterative virtuous circle is conducted to progressively identify “harder” noises and learn better representations. Eventually, in \textbf{Stage 2} (Section~\ref{sec:4.2}), a long-tailed loss, \eg, a balanced loss~\cite{ren2020balanced}, is attached to the clean backbone from Stage 1 to learn a robust classifier. 


\begin{algorithm}[t]
\scriptsize

	\caption{H2E Framework}
	\textbf{Input:} NLT-Dataset $\{(x,y)\}$, $\#$ Iteration $T$, Confidence Threshold $\tau $.
	\begin{algorithmic}[1]

	\State \textbf{Stage0} (Input: \{$\{(x,y)\}$, $\tau $\}) $\to$ Output: \{$\Phi_0(\cdot)$, $f_0(\cdot)$\}  
	
     Initialize backbone $\Phi_0(\cdot)$, linear classifier $f_0(\cdot)$ by Part A in Appendix.
	
	 $\;$
	
		\For {$t=1,2,\ldots T $}
		
		\State \textbf{Stage1} (Input: \{$\{(x,y)\}$, $\Phi_{t-1}(\cdot)$,$f_{t-1}(\cdot)$, $g_{t-1}(\cdot)$\}) $\to$ Output: \{$\Phi_t(\cdot)$, $f_t(\cdot)$, $ g_{t}(\cdot)$\} 
		
		$//$ Learn Noise Identifier.
		
		
		$\;\;\; \{ e_1, e_2, \cdots\}$ generated  multiple environments with Sec.~\ref{sec:4.1}. 
		
		$\;\;\; g_{t}(\cdot)$ $\gets$ $g_{t-1}(\cdot)$ by learning parameters $w$ through IRM with Eq.\eqref{eq.2}.

		
		
          $//$ Easy Noise Removal.
          
          $\{(\tilde{x} ,\tilde{y})\}$ $\gets$ $\{(x,y)\}$ by commensurate Mixup with Eq.\eqref{eq.4}.
          
          $\Phi_{t}(\cdot)$ $\gets$ $\Phi_{t-1}(\cdot)$, $f_{t}(\cdot)$ $\gets$ $f_{t-1}(\cdot)$ by fine-tuning on $\{(\tilde{x} ,\tilde{y})\}$.
		
		\EndFor
	   
	 $\;$ 
	   
    \State \textbf{Stage2} (Input: \{$\{(x,y)\}$, $\Phi_{T}(\cdot)$, $f_{T}(\cdot)$, $g_{T}(\cdot)$ ) $\to$ Output:  updated $f_{T}(\cdot)$ 
    
    $//$ Robust classifier tackling class imbalance.
    
    Update $f_{T}(\cdot)$ by reweighted Balance-softmax from Eq.\eqref{eq.5}.
    
	\end{algorithmic}
	\textbf{Output: } The final robust model $f_T(\Phi_{T}(\cdot))$ .
\label{algo:1}
\end{algorithm}

\subsection{Stage 1: Hard-to-Easy Noise Converter}
\label{sec:4.1}

\noindent\textbf{Input} : An initialized model containing backbone $\Phi(\cdot)$ and projection layer $f(\cdot)$, the training dataset $\{(x,y)\}$.

\noindent\textbf{Output} : A fair noise identifier $g(\cdot)$ invariant to environments, a fine-tuned cleaner backbone $\Phi(\cdot)$ and projection layer $f(\cdot)$.

As we discussed in Section~\ref{sec:3}, the imbalanced $p(z_e | y=c, z_c) \cdot p(y=c)$ turns  ``easy'' noise into ``hard'', since the noise identifier $p(z_c|y=c)$ cannot be disentangled from the context and class bias. To better adapt to the long-tailed classification, the proposed noise identifier combines the previous LWS~\cite{kang2019decoupling} and Logit Adjustment~\cite{menon2020long} classifiers as $g(\cdot) = f(\Phi(\cdot)) - w \cdot\log \pi$, where $\Phi(\cdot)$ is the frozen backbone extracting the image feature; $f(\cdot)$ projects feature vectors to the logit space; ${w}$ is learnable parameters; $\pi$ is the class distribution $p(y)$. However, the above $g(\cdot)$ can only remove the class bias $p(y=c)$ but not the context bias $p(z_e|y=c,z_c)$.

Intuitively, the crux for mitgating context bias $p(z_e|y=c,z_c)$ is to directly eliminate the impact of certain context $z_e$ distribution, making $g(\cdot)$ an invariant identifier by capturing the class-specific attributes $z_c$.
Inspired by Invariant Risk Minimization (IRM)~\cite{arjovsky2019invariant}, we construct a set of environments $\mathcal{E}=\{e_1,e_2,...\}$, ensuring the diverse $p(z_e|y=c,z_c)$ in different environments. Then, IRM essentially regularizes $g(\cdot)$ to be equally optimal across environments with different context-distribution, thus removing the influence of context bias. The objective function of the proposed noise identifier invariant across $\mathcal{E}$ is thus defined as follows:
\begin{equation}
    \label{eq.2}
    \begin{split}
        \min_{g} \; & \sum_{e\in \mathcal{E}} R^e(x,y;f(\Phi(\cdot)),g) \\
        \text{subject to} \; & g \in \arg\min_{g} R^e(x,y;f(\Phi(\cdot)), g)\text{ for all } e\in \mathcal{E},
    \end{split}
\end{equation}
where $R^e(x,y;f(\Phi(\cdot)),g)$ is the risk under environment $e$; $g \in \arg\min_{g} R^e(x,y;M,g)\text{ for all } e\in \mathcal{E}$ means that the invariant identifier $g$ should minimize the risk under all environments simultaneously. The  implementation of IRM loss is in Appendix.
Detailed process of Hard-to-Easy transformation is as below :




\begin{wrapfigure}{r}{5.2cm}
\includegraphics[scale=0.21]{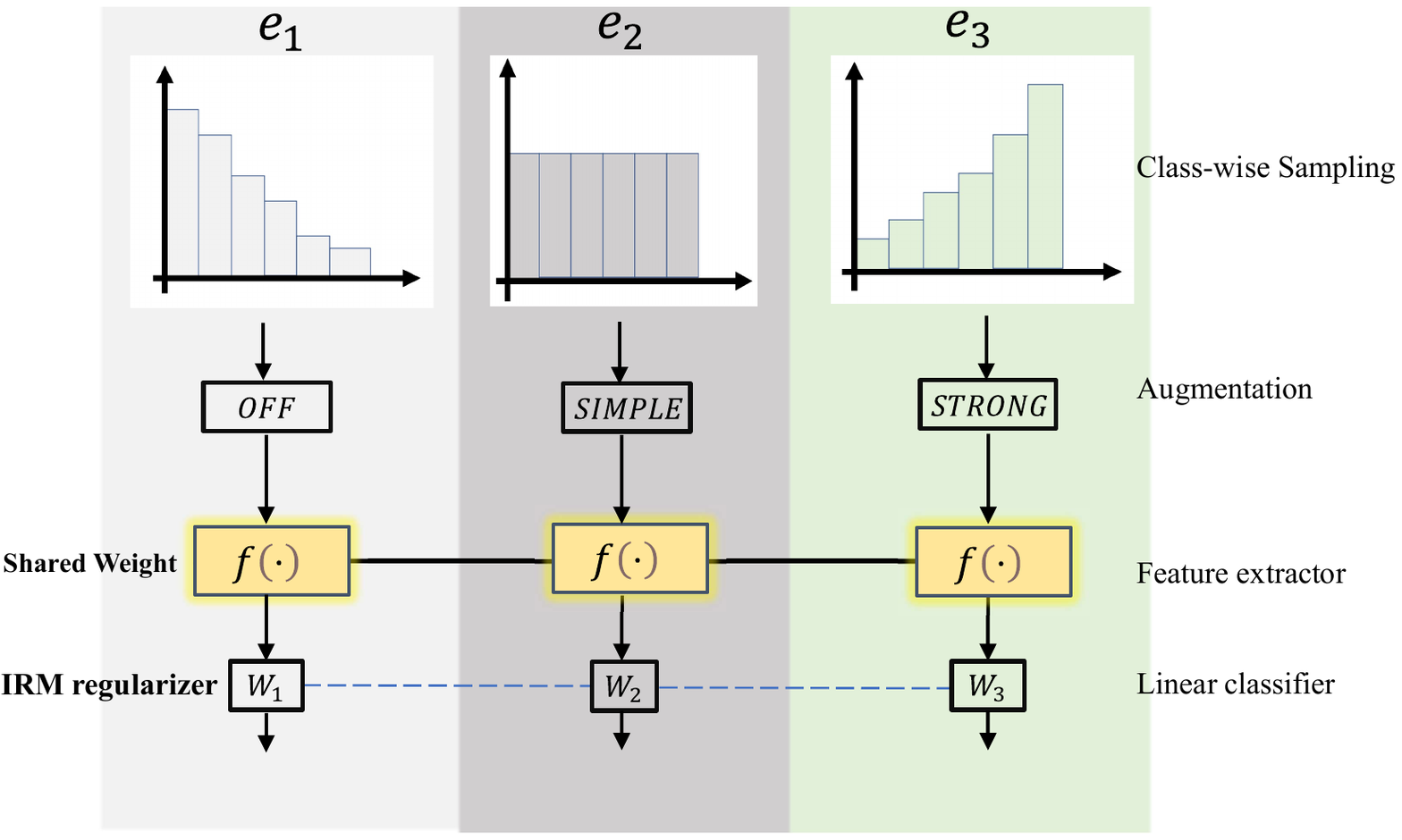}
\caption{\scriptsize{Multi-environment with diverse class and context distributions are built, then an IRM optimization~\cite{arjovsky2019invariant} is applied to obtain an invariant identifier across environments}}
\label{fig:3}
\end{wrapfigure}

\noindent\textbf{Environment Construction.}
A set of diverse environments $\{e_1,e_2,...\}$ are constructed which ensure the variance of $p(z_e|y=c,z_c)$. The criterion of ideal environment construction is the orthogonality of context distribution; however, considering the computation consumption, we only construct three learning environments with classical sampling strategies and provide further ablations of the settings of environment construction in Sec.~\ref{sec:5.4}. As illustrated in Fig.~\ref{fig:3}, each learning environments adopts a different class-wise sampling strategy: 1) the instance-balanced sampler maintains the raw distribution of dataset, 2) the class-balanced sampler ensures the equal probability of being selected for each class, and 3) the class-reversed sampler aims to over-correct the imbalanced $p(y)$ by deliberately picking samples of class $y=c$ with the probability negatively correlated with class size. 
Then, in order to generate diverse distributions of $p(z_e | y=c,z_c)$ to avoid the over-sampling that generates a lot of duplicate samples (especially in tail categories), we adopt different data augmentation methods for each environment: $e_3$  with class-reversed sampler is equipped with the \textbf{``Strong''} augmentation~\cite{cubuk2020randaugment} as it has the most number of duplicate samples, $e_2$ with class-balanced sampler uses \textbf{``Simple''} Random Flip and Resized Crop, as it has less duplicates, and  $e_1$ is without augmentation (\textbf{``OFF''}) as it has no duplicate samples.

\noindent\textbf{Easy Noise Removal.} After obtaining the robust noise identifier $g(\cdot)$, in order to learn a better backbone with less contamination from the noise and prevent those clean images from being mistakenly penalized, we adopt a soft noisy removal strategy that uses Mixup~\cite{zhang2017mixup} to dynamically augment samples according to the confidence generated by the noise identifier $g(\cdot)$. Specifically, we fine-tune $\Phi(\cdot)$ and $f(\cdot)$ with the generation of training pair $(\tilde{x}_{ij}, \tilde{y}_{ij})$ through conducting linear mixture of each two images as follows:
\begin{equation}
\begin{split}
\tilde{x}_{ij} & =\delta_{ij} x_{i}+(1-\delta_{ij}) x_{j},\\
\tilde{y}_{ij} & =\delta_{ij} y_{i}+(1-\delta_{ij}) y_{j},
\end{split}
\label{eq.4}
\end{equation}  
where $x_i$ and $x_j$ are two images with labels $y_i$ and $y_j$, respectively; $\delta_{ij}$ is the de-noise weight in proportion to the confidences $g(x_i)/g(x_j)$. Intuitively, the sample with a higher probability of being noise will have smaller weight in the mixed image $\tilde{x}_{ij}$. Such commensurate Mixup strategy  alleviates the noise memorization effect\cite{li2020gradient} and prevent the overfitting of the already-detected easy noises. Moreover, the long-tailed effect can also be better eliminated by the Mixup, compared with other noise identification methods~\cite{jiang2018mentornet,han2018co,li2020dividemix}.






\noindent\textbf{Iterative Refinement}
As both the noise identifier and the easy noise removal can benefit from the improvement of each other, we introduce an iterative framework to progressively identify ``harder'' noises and learn better representations, refers to $(\Phi_{t-1},f_{t-1},g_{t-1}) \to (\Phi_{t},f_{t},g_{t})$ in Algorithm~\ref{algo:1}.  It's worth noting that the iterative framework needs an initial step to learn a relatively pure feature representation by filtering the ``simplest'' noises, \ie, those samples that both class-specific contents $p(z_c|y=c)$ and context-specific environments $p(z_e|y=c,z_c)$ are obvious outliers of the corresponding class $y=c$. Due to the model memorization effect\cite{li2020gradient}, those ``simplest'' noises can be identified by a commonly adopted warm-up stage when the noisy data haven't affected the learning of generalized patterns yet. The detailed implementation of this step is in Appendix.

\subsection{Stage 2: Robust Classifier}
\label{sec:4.2}
\noindent\textbf{Input} : The last-iteration model $f(\Phi(\cdot))$ in Stage 1, noise identifier $g(\cdot)$, and training dataset $\{(x,y)\}$.

\noindent\textbf{Output} : The final robust model $f(\Phi(\cdot))$.

For the sake of simplicity of symbol, we omit all the subscript in this section.  After obtaining purified representation from the iterative H2E, we assume the network has already modeled the underline $p(x|y=c)$. Therefore, we only need to tackle the class bias $p(y=c)$ in the cleaner data by any existing long-tailed classification algorithms. Without loss of generality, we resort to the balanced softmax loss~\cite{ren2020balanced}, which can be defined as:
$\mathcal{L}(x, y)=-y \log C\!E \left(f                                                     (\Phi(x))+\log \pi\right) $   
, where $C\!E$ denotes the cross-entropy loss; $\pi$ is the distribution of $p(y)$ in the training data; $f(\cdot)$ is the learnable linear classifier initialized from the last-iteration in Stage 1. Noted that the noise identifier $g(\cdot)$ is not directly selected as the initialized classifier since there is a trade-off between the robustness of noise identification and performance of classification.

Besides, to eliminate all the noises detected by $g(\cdot)$, the final robust classifier is thus optimized by re-weighting all samples from the training data according to $\theta(x,y)$ as follows:
\begin{equation}
\mathcal{L}_{overall} = \frac{1}{N} \sum_{(x,y)} \theta(x,y) \cdot \mathcal{L}(x, y),
\label{eq.5}
\end{equation}
where $N$ refers to the batch size, $\theta(x,y)$ is the weight parameter generated by $g_{I}(\cdot)$; $\theta(x,y) = p(y|x)$ when $p(y|x)$ is larger than any other $p(y^{\prime}|x)$ and $\theta(x,y) = \eta$, a hyper-parameter threshold otherwise in order to make noisy samples contribute less to the loss.

\section{Experiments}
\label{sec:5}

\subsection{Benchmarks} 

We constructed three benchmarks for Noisy Long-Tailed (NLT) classification using both synthetic and realistic noise with class imbalance to imitate the real-world dataset at scale. As conventions~\cite{jiang2020beyond}, we call them \textbf{blue} (synthetic) and \textbf{red} (realistic), respectively. Our benchmarks: ImageNet-NLT, Animal10-NLT and Food101-NLT are built on top of three standard image classification dataset : Red Mini-ImageNet~\cite{jiang2020beyond}, Animal-10N~\cite{song2019selfie} and Food-101N~\cite{lee2018cleannet}. 

\noindent\textbf{Dataset Construction}.
During the dataset construction, we adopted the standard rule of first building a balanced but noisy dataset and then transforming them into the long-tailed distribution to simulate the real distribution of noisy labels. To be specific, as for ImageNet-NLT, we augmented the vanilla Mini-ImageNet by adding correct-annotated samples from ImageNet with the same taxonomy. Then we followed the construction of Red Mini-ImageNet to replace $\rho$ proportion of the original training images with noisy images from the web where $\rho$ denotes the noise rate that is uniform across classes. Blue noises in ImageNet-NLT were generated by randomly sampling $\rho$ training images from each class and substituting their labels uniformly drawn from other classes. The above process is not necessary for the construction of Animal10-NLT and Food101-NLT since their original datasets have already contained various real-world noises. After obtaining the balanced but noisy datasets, we simulated the long-tailed distribution in the real-world following the same setting as LDAM~\cite{cao2019ldam} to clip the size of each class: the long-tailed imbalance follows an exponential decay in the number of training samples across different classes. The imbalance ratio $\eta $ denotes the ratio between the size of the maximum and minimum class. 

Before sampling the long-tailed subsets of the original datasets, the balanced Red Mini-ImageNet contains 60,000 images from the original Mini-ImageNet~\cite{sun2019meta} and 54,400 images with incorrect labels collected from the web. Animal-10N is a real-world noisy dataset of human-annotated online images of ten bewildering animals, with 50,000 training and 5,000 testing images in an estimated 8 $ \% $ noise rate. Food-101N is a webly noisy food dataset containing 310,000 images from Google, Yelp, Bing and other search engines using the Food-101~\cite{bossard2014food} taxonomy. Table ~\ref{tab:1} summarizes our benchmarks and further description are in Appendix.

\begin{table}
 \caption{\scriptsize{Overview of three NLT benchmarks with controlled noise level and imbalance ratio}}
\scriptsize
\centering
\begin{tabular}{l | c | c | c | c | c}
\hline
\textbf{Dataset}             & \textbf{\#Class} & \textbf{Train Size} & \textbf{Val Size} & \textbf{Noise Levels(\%)} & \textbf{Imbalance Ratio} \\ \hline
Red ImageNet-NLT    & 100     & 31,817      & 5,000     & 10,20,30      & 0,20            \\
Blue ImageNet-NLT   & 100     & 31,817      & 5,000     & 10,20,30      & 0,20            \\
 \hline 
Food101-NLT         & 101     & 63,460      & 25,000   & $\simeq $8.0        & 20,50,100,200   \\
Animal10-NLT        & 10      & 17,023      & 5,000     & $\simeq $18.4       & 20,50,100,200   \\ 
\hline

\end{tabular}
\label{tab:1}
\end{table}


\subsection{Implementation Details} 
We compared the proposed H2E with previous state-of-the-art methods in both fields of learning with noise and long-tailed classification. Moreover, since noisy long-tailed classification is rarely explored and the number of algorithms designed to fit our setting is small, we further proposed several joint algorithms that combine both long-tailed algorithms and de-noise methods for ablation. 

\noindent\textbf{LT Baselines:} 
1) LWS~\cite{kang2019decoupling} decouples the learning procedure into representation learning and classifier fine-tuning, that re-scales the magnitude of classifier after obtaining the model capable of recognizing all classes; 2) The post-hoc logit adjustment (LA)~\cite{menon2020long} is another widely-used algorithm to compensate the long-tailed distribution by adding a class-dependent offset to each logit; 3) BBN~\cite{zhou2020bbn} uses a framework of Bilateral-Branch network with a cumulative learning strategy; 5) LDAM~\cite{cao2019ldam} is a label-distribution-aware margin loss designed to re-balance the distribution.

\noindent\textbf{De-noise Baselines :}
1) Co-teaching+~\cite{yu2019does} trains two networks then predict first, and selects small-loss data to teach its peer by keeping the data with prediction disagreement only; 2) Nested Co-teaching (N-Coteaching)~\cite{chen2021boosting} conducts adaptive data compression to train two separate networks and is further fine-tuned with Co-teaching (iii). 3) Co-Learning~\cite{tan2021co} further predigests these co-training methods through a shared feature encoder; 4) MentorMix~\cite{jiang2020beyond} minimizes the empirical risk using curriculum learning to overcome both synthetic and realistic web noises; 5) Normalized Loss (NL)~\cite{ma2020normalized} combines passive and active loss to prevent over-fitting to noise labels. 6) Confident Learning (CL)~\cite{northcutt2021confident} is a muti-round learning method which refines the selected set of clean samples by repeating the training round. (7) Two well-known SOTA denoise algorithms JoCoR~\cite{wei2020combating} and DivideMix~\cite{li2020dividemix} are also included.

\noindent\textbf{Joint Baselines:}
1) HAR~\cite{cao2020heteroskedastic} is the first algorithm to tackle the long-tailed distribution with label noises (synthetic ones), that applies a Lipschitz regularizer with varying regularization to deal with noisy and rare examples in a unified way; 2) Co-teaching-WBL (Co-WBL) conducts a temperature weight to offset the tail classes in the procedure of Co-teaching ~\cite{han2018co} and fine-tunes with the balanced softmax loss~\cite{ren2020balanced}; 3) We also intuitively add Re-sampling strategy into the MentorMix~\cite{jiang2020beyond}, denoted as MentorMix-RS, to re-balance before curriculum learning; 4) A distribution-robust loss function ~\cite{cao2019ldam} and a noise-robust loss function~\cite{ma2020normalized} is also combined, denoted as LDAM+NL. 

 \noindent\textbf{Experimental Details.}  
 ResNet-18~\cite{he2016deep} backbone was adopted for all methods in ImageNet-NLT and Animal10-NLT, and ResNet-50~\cite{he2016deep} for Food101-NLT. They were all trained \textit{from scratch} by SGD with weight decay of $1 \times 10^{-4}$ and momentum of 0.9. All models were implemented in PyTorch and on NVIDIA Tesla A100 GPUs for 200 epochs with batch size of 512, except for Co-teaching+~\cite{yu2019does} and Co-teaching-WBL with the batch size of 256. The initial learning rate was set to 0.2 and the default learning rate decay strategy is Cosine Annealing scheduler except for \cite{zhou2020bbn}, \cite{cao2019ldam}, \cite{jiang2020beyond}, which we followed the original setting to apply the multi-step scheduler, and we also maintained the warm-up stage and their backbone variations based on the corresponding papers. It's worth noting that we reported the version of single iteration H2E as well for fair comparison, which is conducted straightforwardly with 200 epochs.  Further experiment on the ablation of iteration is included in Appendix.

\subsection{Main Results} 
\label{sec:5.3}

\begin{figure}[t]
\setlength{\belowdisplayskip}{1pt}
   \begin{minipage}[b]{1.0\linewidth}
   \centerline{\includegraphics[height=45mm, width=115mm]{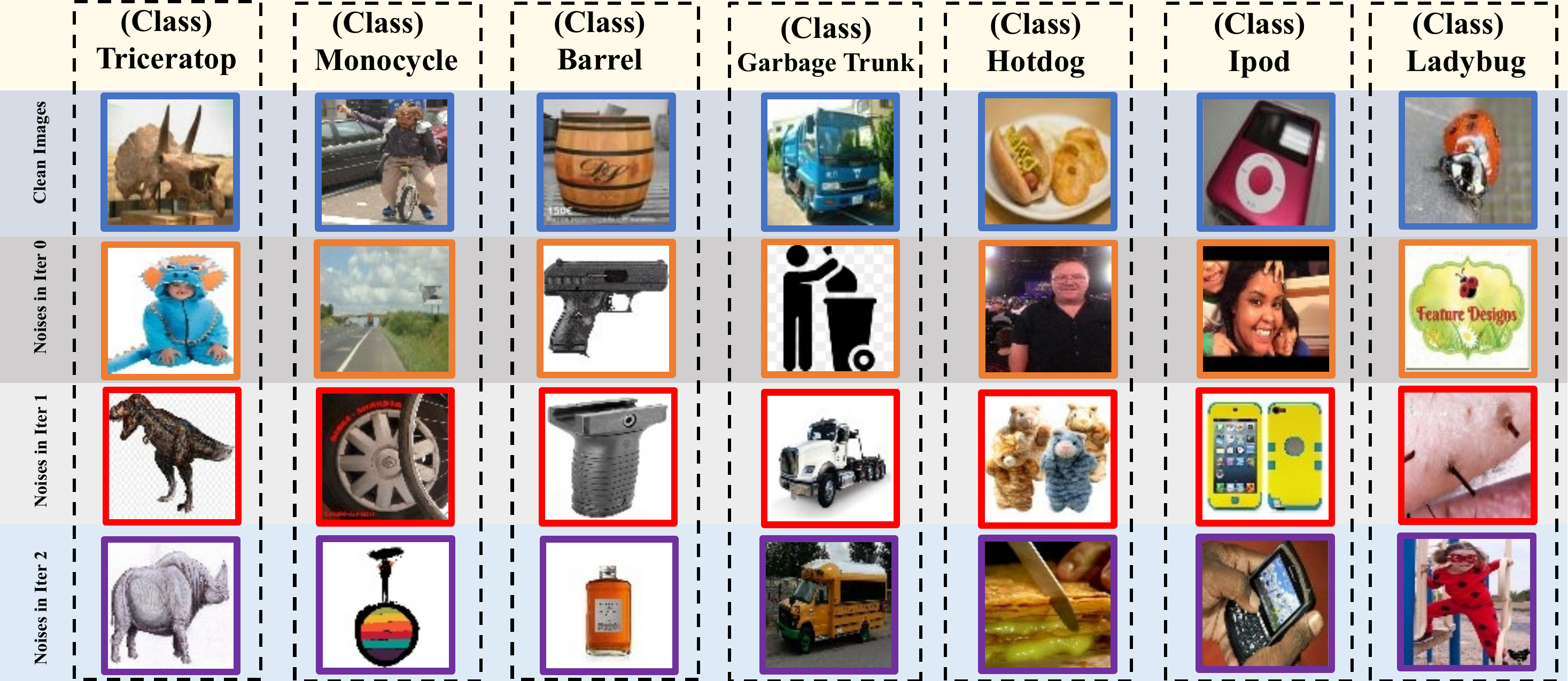}}
   \end{minipage}
   \vspace{-3mm}
   \caption{The example of iterative hard-to-easy transformation on Red ImageNet-NLT, presenting H2E gradually detects harder noises and improve overall robustness }
\label{fig:4}
\end{figure}

\noindent\textbf{Evaluation on ImageNet-NLT}.
We conducted extensive experiments on ImageNet-NLT with three different noise ratios including both synthetic noises and realistic web noises, denoted as blue noises and red noises, respectively, following the setting of Jiang \etal ~\cite{jiang2020beyond}. Fig.~\ref{fig:4} presents that the iterative noise detection can better transfer hard noises into easy ones thus improving the model robustness step by step. We compared our method with several popular LT, de-noise baselines and a few joint baselines were also proposed in our experiments to intuitively combine LT algorithms with de-noise methods. As presented in Table~\ref{tab:1}, the proposed H2E consistently outperforms the baseline methods across different noise rates and noise types (red and blue). In particular, compared with MentorMix~\cite{jiang2020beyond}, which achieves the best performance among selected de-noise methods, H2E improves the test accuracy by 6.1$\%$ on average.

\setlength{\tabcolsep}{2.0mm}{
\begin{table}[t]
\centering
\caption{The evaluation (Top-1 Accuracy\%) on ImageNet-NLT: we reported both blue (synthetic) and red (realistic) noises with three different noise rates: $10\%$, $20\%$, and $30\%$. Experiments demonstrate the effectiveness of the proposed H2E on all settings. The reported H2E-iter has the same number of total epochs with others}

\scalebox{0.90}{

\begin{tabular}{l | l | cc | cc | cc} 

\hline
\hline
\multirow{2}{*}{Category}            & \multirow{2}{*}{Methods} & \multicolumn{2}{l|}{~~~~~~10\% $\rho$
  } & \multicolumn{2}{l|}{~~~~~~20\% $\rho$} & \multicolumn{2}{l}{~~~~~~30\% $\rho$}  \\
                                     &                          & red   & blue                          & red   & blue                          & red   & blue                           \\ 
\hline 
Baseline                             & CE                       & 54.36 & 45.80                          & 50.20  & 40.66                         & 46.90  & 34.80                           \\ 
\hline
\multirow{5}{*}{Denoise
 }  & Co-teaching+~\cite{yu2019does}             & 45.58 & 53.16                         & 44.14 & 49.43                         & 43.16 & 37.47                          \\
                                     & CL~\cite{northcutt2021confident}                       & 52.44 & 48.26                         & 51.42 & 44.23                         & 48.62 & 38.21                          \\
                                     & MentorMix~\cite{jiang2020beyond}                & 59.26 & 54.60                          & 55.18 & 50.20                          & 54.68 & 45.84                          \\
                                     & NL ~\cite{ma2020normalized}                      & 56.36 & 52.48                         & 53.84 & 44.80                          & 51.28 & 39.14
                                     \\
                                     
                                     &Co-learning~\cite{tan2021co}   &50.19  &49.72   & 48.77 &42.65  &44.37  &37.20
                                     \\ 
\hline
\multirow{4}{*}{LT
  } & LWS~\cite{kang2019decoupling}                      & 57.05 & 52.36                         & 53.62 & 44.78                         & 49.15 & 36.54                          \\
                                     & LA ~\cite{menon2020long}                      & 58.92 & 51.34                         & 54.50  & 45.24                         & 51.94 & 37.86                          \\
                                     & BBN~\cite{zhou2020bbn}                      & 57.83 & 52.24                         & 54.88 & 45.76                         & 51.58 & 41.35                          \\
                                     & LDAM~\cite{cao2019ldam}                     & 59.24 & 53.02                         & 55.98 & 46.60                          & 54.38 & 42.76                          \\ 
\hline
\multirow{5}{*}{Joint
  } & HAR \cite{cao2020heteroskedastic}                     & 57.14 & 53.24                         & 54.04 & 47.14                         & 52.13 & 43.92                          \\
                                     & NL+LA                    & 59.80  & 51.88                         & 57.21 & 46.52                         & 53.56 & 37.40                           \\
                                     & Co-WBL          & 61.44 & 54.98                         & 57.62 & 52.40                          & 54.08 & 45.81                          \\
                                     & LDAM+NL                  & 60.06 & 52.90                          & 56.24 & 48.14                         & 54.03 & 43.62                          \\
                                     & MentorMix-RS                & 62.20  & 55.44                         & 56.14 & 52.85                         & 55.91 & 48.27                          \\ 
\hline
\multirow{2}{*}{Ours
}       & H2E                      & 64.86 & 58.12                         & 60.92 & 55.84                        & 58.38 & 51.52                          \\
      & H2E-iter                    & \textbf{65.29} & \textbf{59.42}                          &\textbf{62.12} & \textbf{56.31}                          & \textbf{60.66} & ~~\textbf{52.57 }                         \\

\hline
\hline
\end{tabular}}
\label{tab:2}

\end{table}}

Besides, we can see from Table~\ref{tab:2} that vanilla long-tailed methods outperform de-noise baselines in most lower noisy situations, while their performance gap is narrowed in a higher noise level. Intriguingly, some de-noise methods such as CL~\cite{northcutt2021confident} and Co-teaching+~\cite{yu2019does} are built upon the strong assumption of class balance and highly rely on the \textit{small-loss trick}, so their performance degrades dreadfully in NLT, even worse than Cross-Entropy in many cases. As for the combined methods, we intuitively followed the essence of de-noise and long-tailed algorithms, and proposed MentorMix-RS and Co-teaching-WBL that outperform their counterparts and each individual component in most cases. However, for other strategies such as NL+LA, the improvement is limited and unstable with the contradiction of their re-balance strategies.

\begin{table}
\caption{Evaluations (Top-1 Accuracy$\%$) on Food101-NLT and Animal10-NLT}

\centering
\setlength{\tabcolsep}{2mm}
\scalebox{0.95}{
\begin{tabular}{c|lll|lll} 
\hline \hline
\multicolumn{1}{c|}{Dataset}              & \multicolumn{3}{l|}{~ ~~~~Food101-NLT}      & \multicolumn{3}{l}{~~~~Animal10-NLT}  \\ 
\hline
\multicolumn{1}{c|}{Methods/ $\eta$} & ~~20    & ~~50    & \multicolumn{1}{l|}{~~100} & ~~20    & ~~50    & ~~100                  \\ 
\hline
CE                                          & 57.21 & 49.94 & 44.71                   & 66.10  & 59.94 & 53.02                \\ 
\hline
NL  ~\cite{ma2020normalized}                                        & 60.13 & 53.42 & 46.29                   & 48.20  & 33.46 & 22.08                \\
N-Coteaching~\cite{chen2021boosting}                                & 52.44 & 40.21 & 29.78                   & 57.54 & 41.40 & 39.04                \\
DivideMix~\cite{li2020dividemix}       &69.46   &57.15   &42.80   &72.43 &65.77 &47.60\\
Co-learning~\cite{tan2021co}           &53.76    &45.92   &35.10   &61.70 &52.76 &43.23\\
JoCoR~\cite{wei2020combating}          &49.07    &32.98   &33.49   &51.29 &44.02 &37.19\\

\hline
LDAM~\cite{cao2019ldam}                                        & 61.35 & 59.29 & 48.61                   & 75.40 & 72.82 & \textbf{68.21}                \\
LA ~\cite{menon2020long}                                         & 62.81 & 55.42 & 52.30                   & 69.08 & 67.78 & 61.89                \\
BBN~\cite{zhou2020bbn}                                         & 63.44 & 57.89 & 53.16                   & 72.14 & 70.26 & 60.08                \\
LWS~\cite{kang2019decoupling}                                        & 61.29 & 54.42 & 51.10                   & 71.16 & 69.35 & 62.40                \\ 
\hline
CL +LA                                      & 50.16 & 42.18 & 39.13                   & 54.14 & 46.23 & 41.92                \\
HAR \cite{cao2020heteroskedastic}                                        & 59.95 & 52.45 & 46.12                   & 71.92 & 68.43 & 62.19                \\
Co-teaching-WBL                             & 58.04 & 52.12 & 53.97                   & 72.43 & 71.06 & 66.60                \\ 
\hline
H2E                                         & \textbf{70.35} & \textbf{63.69} & \textbf{58.66}                   & \textbf{77.04} & \textbf{74.94} & 66.58                \\
\hline \hline
\end{tabular}}
\label{tab:3}
\end{table}

Not surprisingly, when we compared results between the red and blue noise settings under the same ratios, all the methods perform much better in red than blue noise except for Co-teaching+ ~\cite{yu2019does}, which applies strong intervention on blue noises. This finding is consistent with Jiang \etal ~\cite{jiang2020beyond}'s conclusion and extends it into a more realistic situation.  We believe the underlying reason behind is that blue noises corrupted by label flipping hurts the representation of the DNN more seriously than those open-set~\cite{wei2021open} and label-dependent red noises, which share more context-specific and class-related attributes. Further analysis of the combination of realistic noise and synthetic noise is given in Appendix.

\noindent\textbf{Evaluation on Animal10-NLT and Food101-NLT}.
We further investigated the performance of H2E and other methods in Animal10-NLT and Food101-NLT with various imbalance ratios $\eta\in\{10,20,50\}$. As shown in Table~\ref{tab:3}, our method retains the most robust performance and outperforms other approaches in most cases as the imbalance ratio increase while most de-noise methods~\cite{chen2021boosting,ma2020normalized} suffer from class imbalance and even perform worse than the Cross-Entropy. Long-tailed methods~\cite{zhou2020bbn,cao2019ldam} perform much better than de-noise methods in Animal10-NLT, attributes to the relatively low noise rate (estimated as 8$\%$) and their specific design on network structures, e.g. cosine classifier in LDAM~\cite{cao2019ldam} and extra blocks in BBN ~\cite{zhou2020bbn}. Note that H2E is still comparable with state-of-art de-noise algorithms in a strictly balance training set, with 85.1$\%$ test accuracy in Animal-10N~\cite{song2019selfie} and  73.4$\%$ test accuracy in Food-101N~\cite{lee2018cleannet} from scratch.

\subsection{Ablation Studies and Further Analysis} 

\label{sec:5.4}

\vspace{-1mm}

\noindent\textbf{Q1: }\emph{\textbf{Why H2E outperforms other methods in NLT?}} To better diagnose the improvement of H2E, we followed ~\cite{liu2019large} and further recorded test accuracy and the precision of noise identification on three splits of classes: Many-shot(the top 25$\%$), Medium-shot(the middle 50$\%$) and Few-shot(the last 25$\%$).

\begin{figure}

   \begin{minipage}[b]{1.0\linewidth}
   \centerline{\includegraphics[height=35mm, width=110mm ]{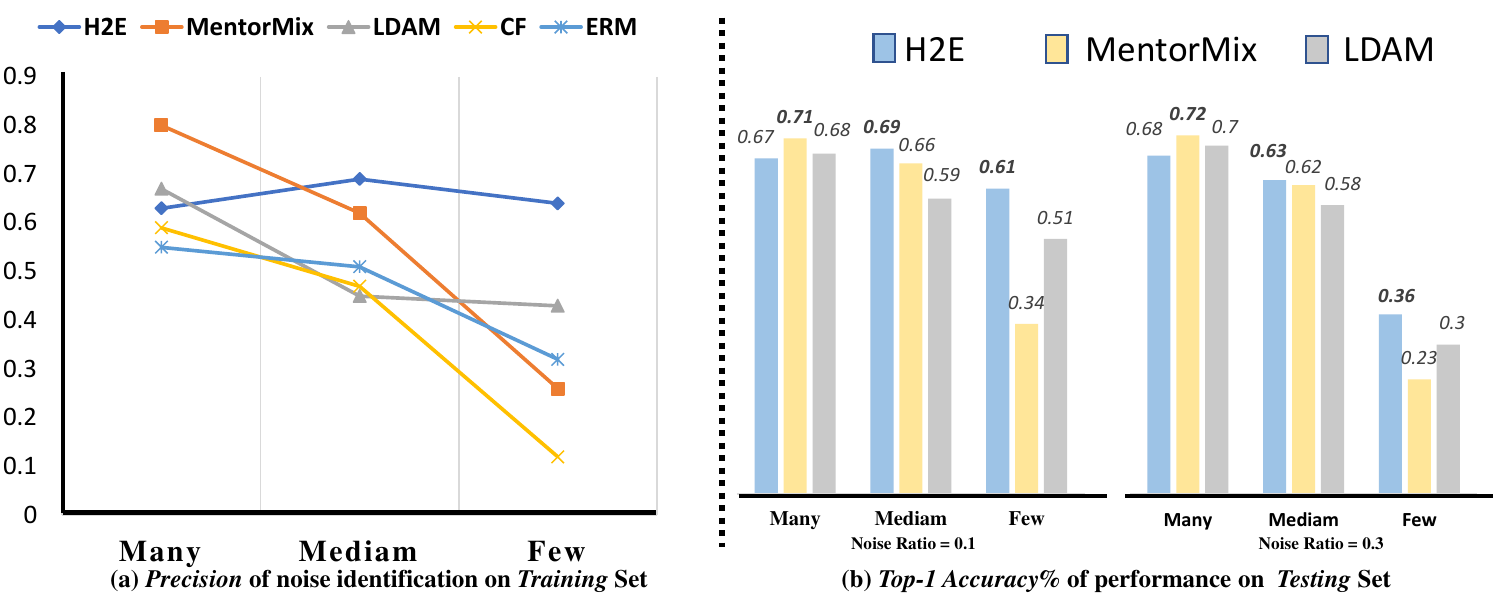}}
   \end{minipage}
   \vspace{-3mm}
   \caption{\scriptsize{(a) Evaluation(Precision) of noise   identification   capability   on Blue  ImageNet-NLT. The  proposed  H2E  indeed significantly improves the Few-shot(tail)  categories  by  better  identifying hard noises. (b)Evaluations (Top-1 Accuracy\%) on Red ImageNet-NLT. We compare test accuracy in Many, Medium and Few shots among different methods}}
\label{fig:5}
\end{figure}

\noindent\textbf{A1: } Specifically in Fig.~\ref{fig:5}(b), the proposed H2E surpasses MentorMix\cite{jiang2020beyond} and LDAM\cite{cao2019ldam} in few-shot by 20 $\%$ and 8$\%$ on average, which concretely demonstrates the robustness of H2E under imbalance distribution. It is clear from  Fig.~\ref{fig:5}(a) that considering the precision of noise detection, H2E outperforms all of the selected methods in tail classes, which highlights its power to identify hard noises. From these two aspects, we could give a conclusion: the higher performance of H2E indeed attributes to its comparatively better hard noise identification capability and less hurt on correct-annotated but rare samples, especially on tail classes.

\begin{table}
        \begin{minipage}[t]{0.5\textwidth}
            \centering
            \makeatletter\def\@captype{table}\makeatother\caption{\scriptsize{Ablation studies of \textbf{env}}}
\scriptsize
 \resizebox{0.9\textwidth}{!}{
\begin{tabular}{ll|ll|ll|ll}
\hline
\multicolumn{2}{l|}{\multirow{2}{*}{Settings/ $\rho$}} & \multicolumn{2}{l|}{\multirow{2}{*}{~~~~10\% }} & \multicolumn{2}{l|}{\multirow{2}{*}{~~~~20\% }} & \multicolumn{2}{l}{\multirow{2}{*}{~~~~30\% }} \\
\multicolumn{2}{l|}{}                          & \multicolumn{2}{l|}{}                                   & \multicolumn{2}{l|}{}                                   & \multicolumn{2}{l}{}                                   \\
\multicolumn{2}{l|}{}                          & \multicolumn{2}{l|}{}                                   & \multicolumn{2}{l|}{}                                   & \multicolumn{2}{l}{}                                   \\

\#env                   & Aug                  & red                        & blue                       & red                        & blue                       & red                        & blue                      \\ \hline
~~~~2                       &                     & 61.40                      & 54.68                      & 57.78                      & 55.18                      & 55.78                      & 48.22                     \\
~~~~2                       & ~$\checkmark$                    & 62.15                      & 55.70                      & 59.66                      & 55.38                      & 56.02                      & 49.10                     \\
~~~~3                       &                     & 62.79                      & 57.40                      & 60.64                      & 55.06                      & 57.14                      & 49.38                     \\
~~~~3                       & ~$\checkmark$                    & 64.86                      & 58.12                      & 60.92                      & 55.84                      & 58.38                      & 51.52                     \\
~~~~4                       &                     & 62.49                      & 55.78                      & 60.18                      & 55.40                      & 56.22                      & 49.30                     \\
~~~~4                       & ~$\checkmark$                    & 65.38                      & 56.52                      & 60.42                      & 55.96                      & 57.56                      & 49.78                     \\ 
\hline

\end{tabular}}
\label{tab:4}
        \end{minipage}
        \begin{minipage}[t]{0.5\textwidth}
        \centering
        \makeatletter\def\@captype{table}\makeatother\caption{\scriptsize{Effectiveness for each component }}
\resizebox{0.9\textwidth}{!}{
\begin{tabular}{ll|ll|ll|ll}
\hline
\multicolumn{2}{l|}{\multirow{3}{*}{Component/ $\rho$}} & \multicolumn{2}{c|}{\multirow{3}{*}{10\%}} & \multicolumn{2}{c|}{\multirow{3}{*}{20\%}} & \multicolumn{2}{c}{\multirow{3}{*}{30\%}} \\
\multicolumn{2}{l|}{}                                                & \multicolumn{2}{c|}{}                      & \multicolumn{2}{c|}{}                      & \multicolumn{2}{c}{}                      \\
\multicolumn{2}{l|}{}                                                & \multicolumn{2}{c|}{}                      & \multicolumn{2}{c|}{}                      & \multicolumn{2}{c}{}                      \\

Stage1                            & Stage2                           & red                  & blue                & red                  & blue                & red                 & blue                \\ \hline
CF                                & CE+RW                            & 56.54                & 48.40               & 51.29                & 43.60               & 48.09               & 38.77               \\
CF                                & H2E                              & 63.90                & 56.10               & 60.08                & 54.76               & 57.80               & 49.30               \\
LDAM                              & H2E                              & 61.94                & 56.42               & 57.10                & 55.69               & 54.31               & 48.46               \\
H2E                               & ERM+RW                           & 60.20                & 57.02               & 58.54                & 50.94 
& 56.27               & 46.24               \\ 
\hline

\end{tabular}}
\label{tab:5}
        \end{minipage}
    \end{table}

\noindent\textbf{Q2: }\emph{\textbf{What impact performance of H2E considering environment construction? }} We conducted two ablation experiments on ImageNet-NLT: one is to analyze the number of environments and the other is to unify the augmentation strategies in each environment to so-called ``OFF'' augmentation.

\noindent\textbf{A2: } From Table~\ref{tab:4}, we found out that the overall improvement is converged to the number of environments when $e>2$ ; Moreover, H2E will averagely degrade by \textit{1.14$\%$} without handling the duplication in tail classes , \ie, not augmentations, but it still largely outperforms other baselines in most settings comparing with the result in Table.~\ref{tab:2}, which shows the power of the proposed H2E.

\noindent\textbf{Q3: }\emph{\textbf{How effective is each individual component in H2E?}} In Table~\ref{tab:4}, we replaced and modified each individual stage in H2E with other feasible methods to examine the effectiveness of the each stage. 

\noindent\textbf{A3: } Considering a muti-stage framework, substituting any part of H2E caused the performance dropping to some extent. In detail, if we replace one component with other baselines, the Top-1 Accuracy will averagely degrade by \textit{2.81\%}.

\noindent\textbf{Q4: } \emph{\textbf{What's the justification of augmentation strategies in environment construction?}}

\noindent\textbf{A4: }(1) All methods in Section~\ref{sec:5} contains the so-called strong augmentation in the stage of data prepossessing for fair comparison, \textbf{so we don't take any unfair advantage}. (2) Different augmentations are introduced only to construct environments with context-wise distribution shift. It's directly derived from our formulation, so IRM can focus on class-specific attributes, making it easier to converge and avoid both class and context bias.


\section{Conclusion}
\label{sec:7}
We presented a novel noisy learning algorithm, Hard-to-Easy (H2E) for Noisy Long-Tailed Classification (NLT). We motivated from the observation that the tail class confidence boundary between clean and noisy samples are not clear, rendering conventional noise identification methods ineffective. Our analysis shows that it is because the class and context imbalance in long-tailed data that turn the ``easy'' noises into ``hard'' ones. The highlight of H2E is that it learns a robust noise identifier invariant to the class and context environmental changes. On three newly proposed NLT benchmarks: ImageNet-NLT, Animal10-NLT, and Food101-NLT, we demonstrated that H2E significantly outperforms existing de-noise methods, which do not take the imbalance into account. In future, we will conduct further analysis on NLT settings and more effective environment-invariant learning algorithms.


\noindent\textbf{Acknowledgements}: This research is supported by the National Research Foundation, Singapore under its AI Singapore Programme (AISG Award No: AISG2-RP-2021-022), Alibaba-NTU Singapore
Joint Research Institute (JRI), and the Agency for Science, Technology AND Research (A*STAR).

\appendix

\section{Implement details}
\subsection{Initializing Stage}
The hard noise identifier phase and easy noise removal proceed iteratively. The detailed implementation of the initializing step will be introduced here. In a general way, it is conducted before the noise removal stage which utilizes the model memorization effect~\cite{li2020gradient}. Li \etal. proved the distance
\begin{equation}
\left\|W_{t}-W_{0}\right\|_{F} \lesssim\left(\sqrt{K}+\left(K^{2} \epsilon_{0} /\|C\|^{2}\right) t\right)    
\end{equation}
from the initial weight $W_{0}$ to current weight $W_{t}$ on a unit Euclidean ball assuming distinguishable samples,where K denotes the scales of clusters, and C is $\epsilon_{0}$-neighborhood cluster centers. It demonstrates that DNNs tend to learn simple and generalized patterns in the first step,then over-fit to noisy patterns from easy to hard.

We use the preliminary network trained in the warm-up stage to extract features $\nu$ in each category and construct a cosine similarity matrix $\mathbf{M} \in \mathbb{R}^{n \times n}$,where $M_{i j}=\frac{\nu \left(\mathbf{x}_{i}\right)^{T} \nu\left(\mathbf{x}_{j}\right)}{\left\|\nu \left(\mathbf{x}_{i}\right)\right\|_{2}\left\|\nu\left(\mathbf{x}_{j}\right)\right\|_{2}}$ measures the similarity between images.

\begin{table}
\centering
\caption{\small{Evaluations (Top-1 Accuracy\%) on Food101N and Animal10N under balanced class distributions.}}
\begin{tabular}{l|ll} 
\hline

\begin{tabular}[c]{@{}l@{}}Methods\\\end{tabular} & \begin{tabular}[c]{@{}l@{}}Animal-10N\\\end{tabular} & \begin{tabular}[c]{@{}l@{}}Food-101N\\\end{tabular}  \\ 
\hline\hline
CE                                                & ~~81.28                                                & ~~69.42                                                \\
NL~\cite{ma2020normalized}                                                & ~~83.24                                                & ~~69.91                                                \\
N-Coteaching~\cite{chen2021boosting}                                      & ~~84.90                                                 & ~~65.72                                                \\
MentorMix~\cite{jiang2020beyond}                                         & ~~84.10                                                 & ~~73.58                                                \\
HAR~\cite{cao2020heteroskedastic}                                              & ~~81.94                                                & ~~71.76                                                \\
DivideMix~\cite{li2020dividemix}                                         & \textcolor[rgb]{0,0,1}{~~85.72}                                                & \textcolor[rgb]{0,0,1}{~~75.83}                                                \\
Co-teaching+~\cite{yu2019does}                                      & ~~83.66                                                & ~~72.97                                                \\
H2E                                               & ~~85.10                                                 & ~~73.34                                                \\
\hline
\end{tabular}

\label{atab:1}
\end{table}

We define the density $\rho_{i}=\frac{1}{\left \| D_{g} \right \| } \sum_{j=1}^{\left \| D_{g} \right \|} \operatorname{M_{i j}}$ for each image in category $D_{g}$. Since the image with less $p$ has more similar images around them, we could detect and give initial weight parameters $W_{D}$ in instance level based on the sequence with the above density. We control this procedure and normalize the weight parameter in both head and tail classes,t hus wouldn't meet self-confirmation bias caused by the imbalanced distribution.

 \begin{figure}
   \begin{minipage}[b]{1.0\linewidth}
   \centerline{\includegraphics[scale=0.53]{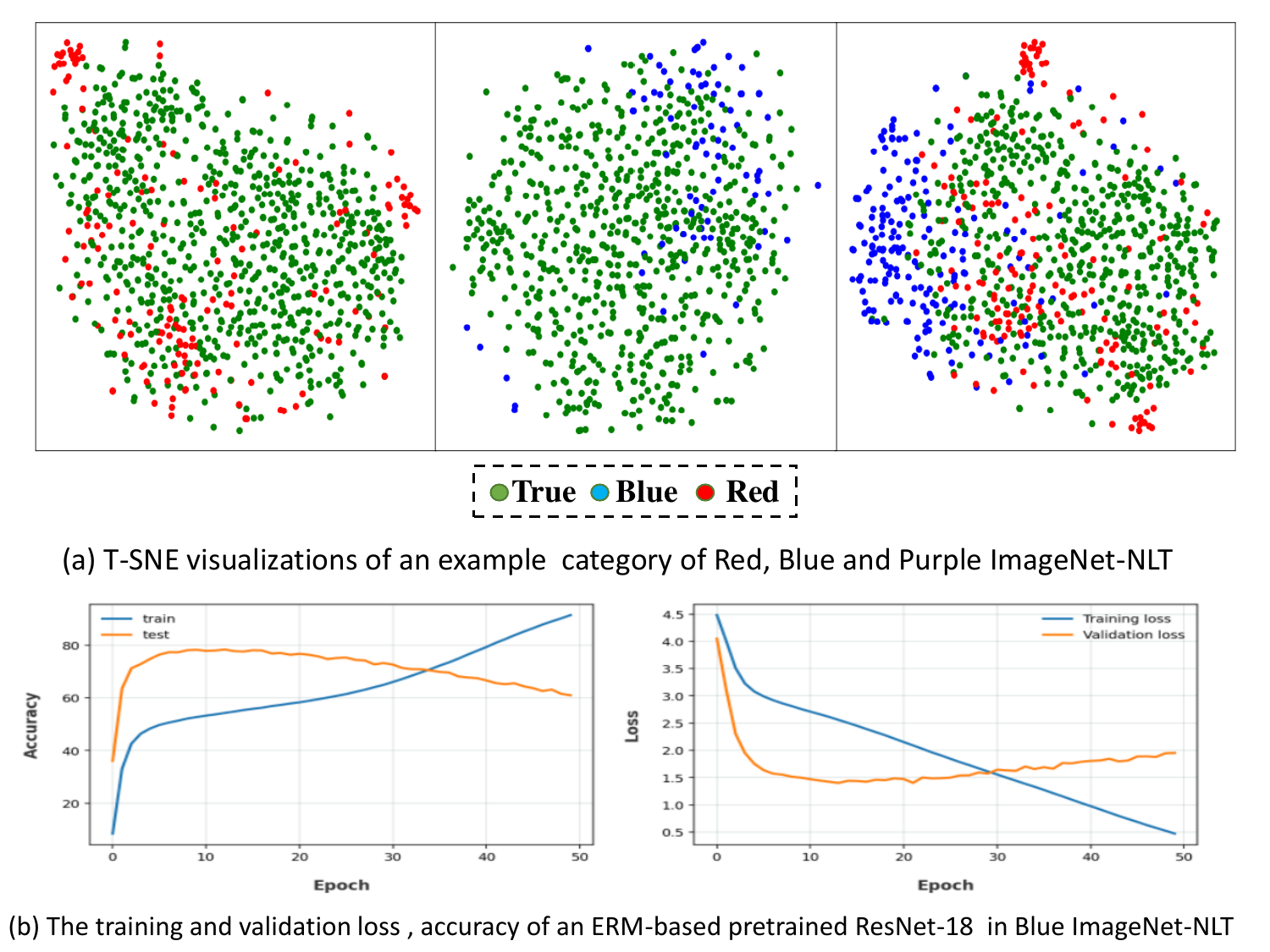}}
   \end{minipage}
   \caption{ (a) The T-SNE~\cite{van2008visualizing} visualization of a certain category in Red, Blue and Purple ImageNet-NLT indicates the distinct patterns between synthetic and realistic noises : realistic noises share more cluster effect and severe confusion with true samples. Figure 3. (b) shows the corruption brought by blue noise degrades the performance of DNN by full over-fitting on mislabelled samples.  }
\label{afig:3}
\end{figure}

\section{Extra experiment}
\subsection{Additional Results on balanced noisy Dataset} 
Although our proposed H2E is particularly designed under both longtailed and noisy datasets, it could still work well on balanced and noisy datasets. In the extra experiment, we just construct one environment and use the balanced softmax loss~\cite{ren2020balanced} to substitute the IRM loss. The implement details are the same as before. Table~\ref{atab:1} gives competitive results compared to various state-of-art denoise algorithms. It demonstrates that without the strong assumption of small loss trick and frequent reweighting (For instance, Co-teaching~\cite{han2018co} samples its small-loss instances as the useful knowledge and teaches it to its peer network for future training.), H2E framework could still show strong robustness when learning with noise on balanced datasets.

\subsection{Additional Results on higher imbalance ratio} 
We further apply our method to the setups with higher imbalance ratio. For instance, in Animal10-NLT with imbalance ratio 200, H2E outperforms CE, MentorMix and BBN by \textit{13.42\%} , \textit{7.12\%} and \textit{2.35\%} respectively.

\subsection{Additional Results on Purple ImageNet-NLT} 
In the main manuscript, we focused on adding one type of noise (synthetic or realistic) and presented its performance for comparison. It is interesting to discuss the results under the longtailed dataset with compound noise, thus we constructed Purple ImageNet-NLT and compared H2E with previous state-of-the-art methods under this new setting. From Table~\ref{atab:2}, our method consistently retains the most robust performance and out-performs other approaches in most cases. This further supports that our proposed framework can adapt to complex noise conditions.

\begin{figure}
   \begin{minipage}[b]{1.0\linewidth}
   \centerline{\includegraphics[scale=0.38]{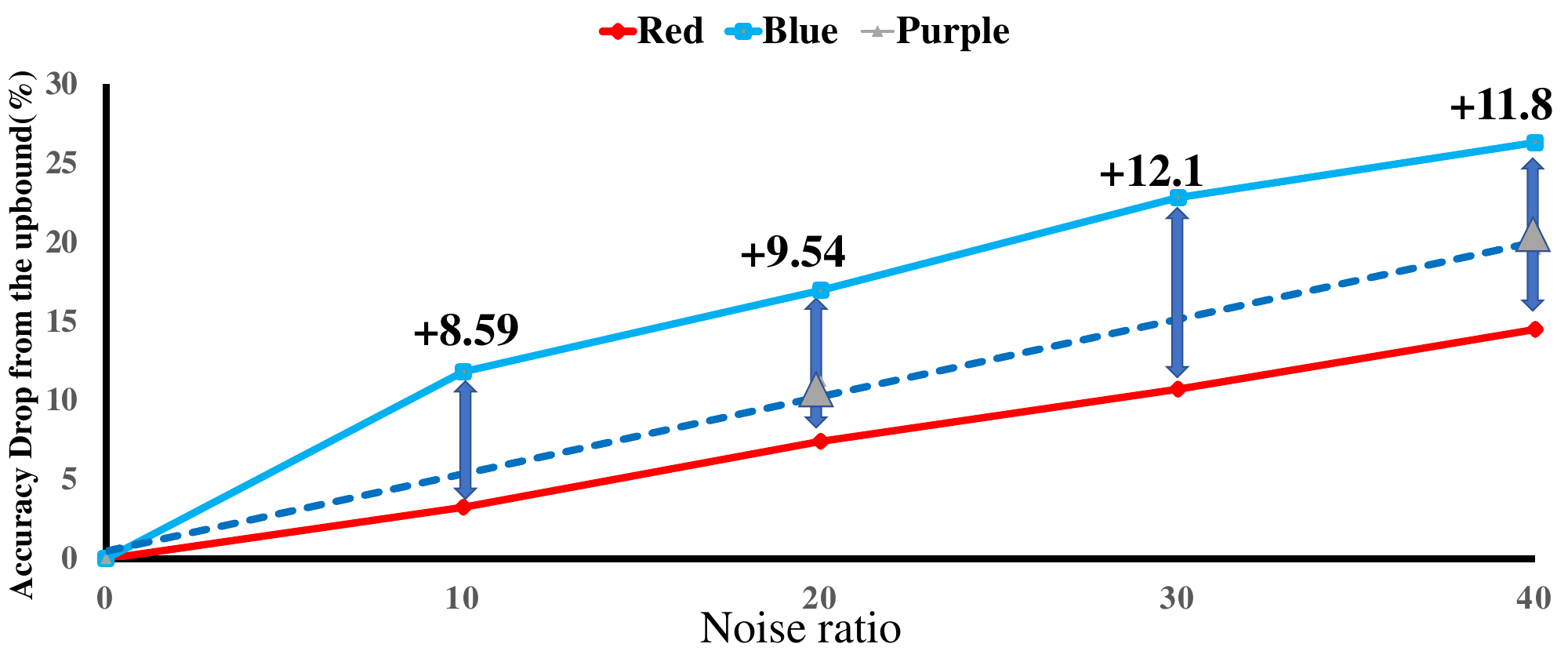}}
   \end{minipage}
   \caption{Performance drop of a ERM-based DNN from the up-bound accuracy in the clean setting with the increase of noise ratios. }
\label{afig:2}
\end{figure}

\subsection{Ablation of Iterative Pattern} 
\begin{figure}
   \begin{minipage}[b]{1.0\linewidth}
   \centerline{\includegraphics[scale=0.38]{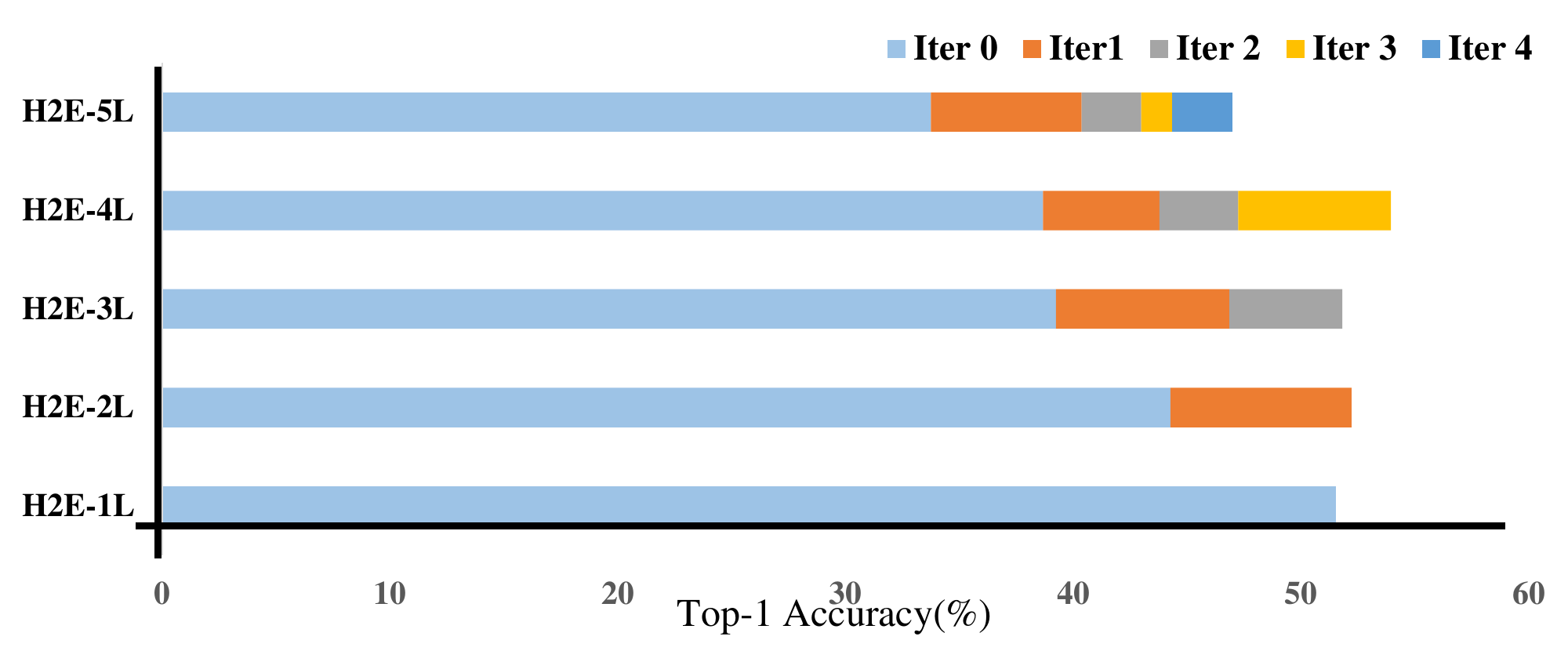}}
   \end{minipage}
   \caption{The ablation of iterative patterns. We reported the proposed H2E with different iteration numbers $n = 1,2,3,4,5$, where each iteration has $Total-Epoch / n$ epochs. The improvement of each iteration is presented with different colors.}
\label{afig:1}
\end{figure}

Sample reweighting is necessary and inevitable in noise sample selection: Some methods~\cite{han2018co,chen2021boosting,yu2019does} conduct the reweighting schedule in frequently. For example, Co-teaching~\cite{han2018co} teaches its peer network with small-loss samples in each mini-batch for future training. Others~\cite{li2020dividemix,jiang2020beyond} apply this procedure as segmentation of clean and noisy after several steps. As an illustration, DivideMix~\cite{li2020dividemix} first train several epochs with confident penalty and then conduct Gaussian mixture model on the loss from the former step to distinguish clean and noisy samples. We believe that the reason why the latter outperforms the former in our settings is largely because frequent sample reweighting in the early stage will extensively degrade the model representation capability in tail classes. The frequency of reweighting procedure should be considered and if we fixed the total number of epochs, there is a trade-off between the average training epochs in each iteration and the total number of loops. We analyzed the effect of the quantity of iterations $n$ in Fig.~\ref{afig:1}, which states clearly that the performance of H2E is relatively stable in certain range ($n = 1,2,3,4)$. However, when the number of epochs per iteration becomes much smaller, the overall performance degrades step by step, which is mainly caused by fact that the 
volatile oscillations of the model with few training epochs cannot support the hard noise identification stage to be better constrained and play a role. The detailed threshold of  the quantity of iterations $n$ changes depending on different settings.

\begin{table}
\centering

\caption{The evaluation (Top-1 Accuracy\%) on Purple ImageNet-NLT: we reported purple noises with two different noise rates: $20\%$, and $40\%$, where red and blue noise has the same proportion. Experiments demonstrate the effectiveness of the proposed H2E on all settings. The reported H2E-iter has the same number of total epochs with others.}

\begin{tabular}{l|l|ll} 
\hline
\multicolumn{1}{l}{Category}         & \multicolumn{1}{l}{Methods} & 20\% noise rate & 40\% noise rate  \\ 
\hline\hline
Baseline                             & CE                          & 46.42           & 38.08            \\ 
\hline
\multirow{4}{*}{Denoise
  Baseline}  & Co-teaching+~\cite{yu2019does}                 & 41.65           & 38.84            \\
                                     & CL                          & 48.49           & 40.14            \\
                                     & MentorMix~\cite{jiang2020beyond}                   & 52.94           & 43.57            \\
                                     & NL~\cite{ma2020normalized}                          & 50.18           & 41.22            \\ 
\hline
\multirow{4}{*}{Longtail
  Baseline} & LWS~\cite{kang2019decoupling}                         & 48.04           & 40.98            \\
                                     & LA~\cite{menon2020long}                           & 52.30            & 42.38            \\
                                     & BBN~\cite{zhou2020bbn}                         & 48.36           & 41.42            \\
                                     & LDAM~\cite{cao2019ldam}                           & 50.42           & 38.92            \\ 
\hline
\multirow{5}{*}{Combined
  Baseline} & HAR~\cite{cao2020heteroskedastic}                         & 49.77           & 38.63            \\
                                     & NL+LA                       & 51.33           & 42.47            \\
                                     & Co-teaching-WBL             & 54.76           & 43.61            \\
                                     & LDAM+NL                     & 53.21           & 41.09            \\
                                     & MentorMix-RS                & 54.82           & 45.14            \\ 
\hline
\multirow{2}{*}{Our
  methods}       & H2E                         & \textbf{59.94}                & \textbf{50.64}                   \\
                                     & H2E-iter                    &  \textbf{61.78}                 &  \textbf{52.22}                  \\
\hline
\end{tabular}

\label{atab:2}
\end{table}

\section{Discussion}
Jiang \etal~\cite{jiang2020beyond} found that DNNS generalize better on red noise and reported the comparison of performance drop from the peak accuracy at different noise levels in blue and red noise settings. They hypothesized DNNS are more robust to web labels since they are more relevant ,in our words, sharing more context-specific attributes, to the clean training samples. We further proved this hypothesis with the following observation and analysis.

\begin{figure}
   \begin{minipage}[b]{1.0\linewidth}
   \centerline{\includegraphics[width=150mm]{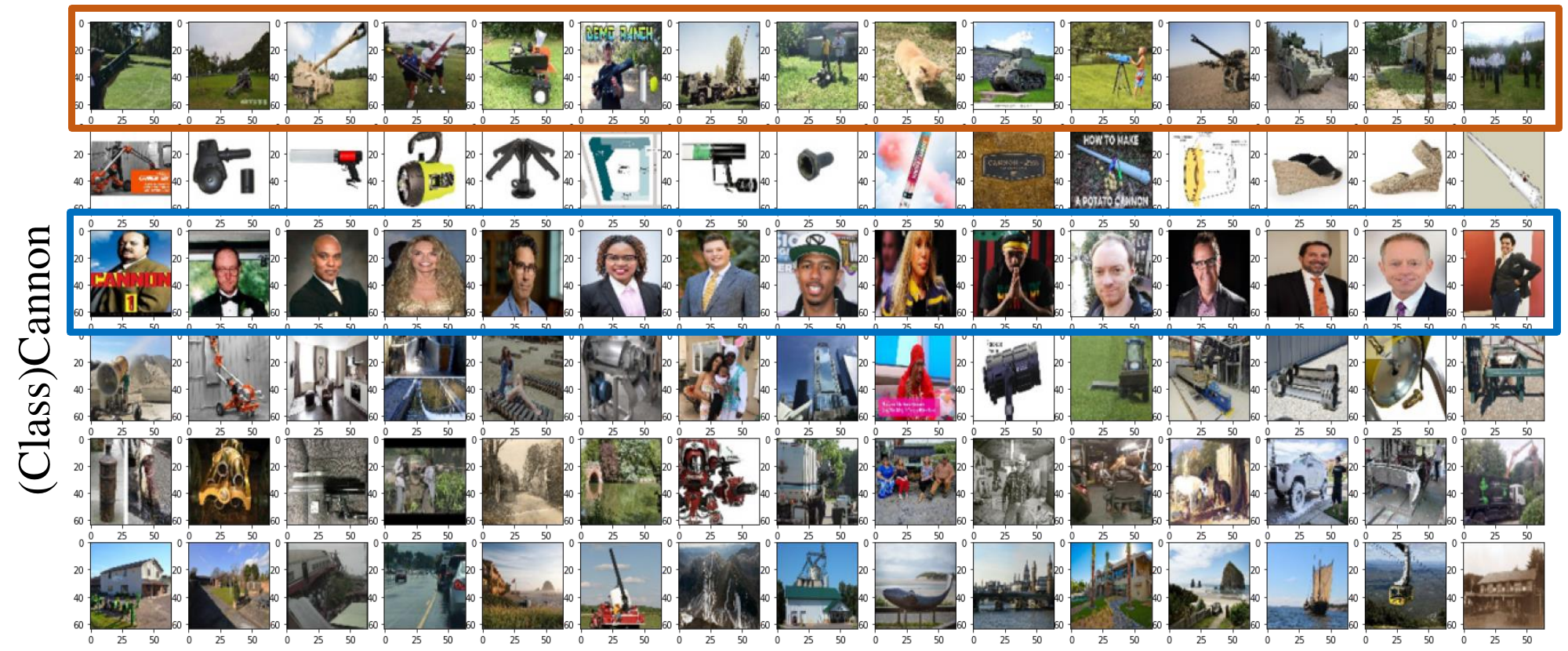}}
   \end{minipage}
   \caption{\normalsize{Visualization of red noise clustering results in class 'Cannon'. }}
\label{afig:5}
\end{figure}

\begin{figure}
   \begin{minipage}[b]{1.0\linewidth}
   \centerline{\includegraphics[width=150mm]{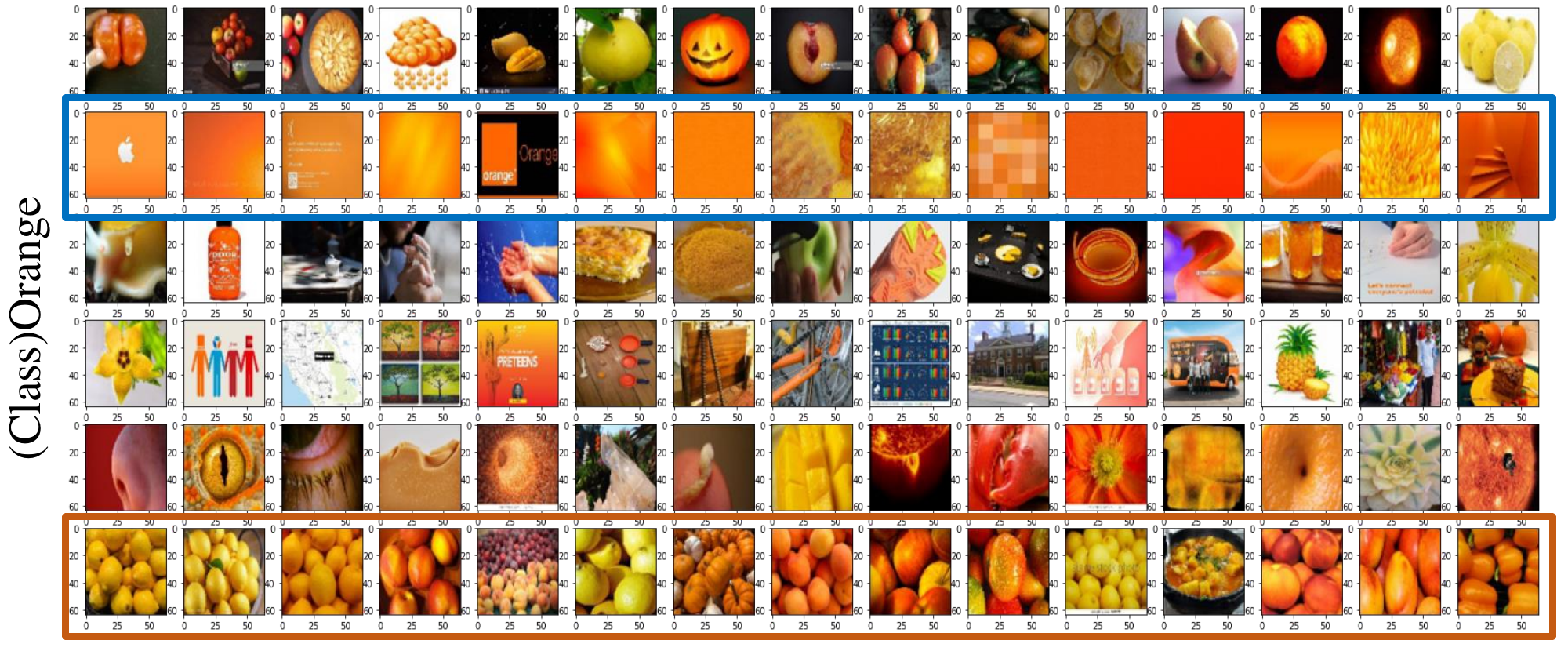}}
   \end{minipage}
   \caption{\normalsize{Visualization of red noise clustering results in class 'Orange'. }}
\label{afig:4}
\end{figure}

\noindent\textbf{DNNS perform much better under red noise}.
Jiang \etal~\cite{jiang2020beyond} noticed the generalization performance of DNNS drops sharply with the ratio of noisy samples increases on blue noise while has relatively smaller difference on red noise. We first confirm Jiang \etal 's conclusion in Fig.~\ref{afig:2}, where with the same noise ratio, the performance drop of DNNS training from scratch is considerably smaller in Red ImageNet-NLT than Blue ImageNet-NLT. However, it needs to be noted that the difference of Top-1 accuracy in this two settings is stable with the increase of noise ratio, which is inconsistent with  Jiang \etal 's finding in the balanced Red and Blue Mini-ImageNet. We consider this disagreement is mainly attribute to the fact that comparing with its counterpart in a balanced setting, red noise could heavily corrupt the tail classes by degrading the diversity of correct-annotated samples.

\noindent\textbf{Red noise presents more cluster effect}.
Jiang \etal~\cite{jiang2020beyond} collected red noise retrieved by Google image search from text-to-image and image-to-image search, which resulted in the fact that part of the red noise contained semantic confusion. For instance in Fig. ~\ref{afig:4} and Fig. ~\ref{afig:5}, for class \textit{Orange}: Fruit of various citrus species in the family Rutaceae, plenty of pictures with orange color are selected from web ; for class \textit{Cannon}:  a large-caliber gun classified as a type of artillery, several artillery commanders and pop singers are picked by Google. In a word, semantic confusion generate red noise and meanwhile cause relatively more cluster effect, which may corrupt some noise identification strategies based on Self-supervised Learning~\cite{karthik2021learning} and Clustering~\cite{wei2021robust}.

\noindent\textbf{Red noise is harder to identify but degrades less}.
We conducted a pretrained ResNet-50 as the feature extractor and gave the T-SNE~\cite{van2008visualizing} visualization of feature representation space in class \textit{prayer rug} with red, blue and purple noise.  Fig.~\ref{fig:3} (a) shows that red noises appear in pairs and confuses heavily with true samples, which makes part of them harder to identify while blue noises present scattered distribution and have clearer boundary with true samples. Here comes the question : \textit{Since the detection of red noises is much difficult than the blue ones, why most of the denoise methods could perform better in red noise settings?} From observation and analysis, two reasons are given: (1) In Fig.~\ref{afig:3} (b), we found that DNN easily fits blue noises(random labels) and causes performance degradation even in the last epochs with relatively robust representation capability, while generalize much better with red noises and maintain stable performance in the last stage with the increase of epochs. (2) Blue noises are generated by random label flipping, which indicates that they have corresponding true labels and classes included in the training set, while most of Red noises from web 
aren't affiliated with any class in the training set. The different attribute of closed-set and open-set led to their different corruption on DNNS to certain extent.

\clearpage
%
%
\bibliographystyle{splncs04}
\bibliography{egbib}

\begin{thebibliography}{10}
\providecommand{\url}[1]{\texttt{#1}}
\providecommand{\urlprefix}{URL }
\providecommand{\doi}[1]{https://doi.org/#1}

\bibitem{amid2019robust}
Amid, E., Warmuth, M.K., Anil, R., Koren, T.: Robust bi-tempered logistic loss based on bregman divergences. arXiv preprint arXiv:1906.03361  (2019)

\bibitem{arazo2019unsupervised}
Arazo, E., Ortego, D., Albert, P., O’Connor, N., McGuinness, K.: Unsupervised label noise modeling and loss correction. In: International Conference on Machine Learning. pp. 312--321. PMLR (2019)

\bibitem{arjovsky2019invariant}
Arjovsky, M., Bottou, L., Gulrajani, I., Lopez-Paz, D.: Invariant risk minimization. arXiv preprint arXiv:1907.02893  (2019)

\bibitem{bernardo2009bayesian}
Bernardo, J.M., Smith, A.F.: Bayesian theory, vol.~405. John Wiley \& Sons (2009)

\bibitem{bossard2014food}
Bossard, L., Guillaumin, M., Van~Gool, L.: Food-101--mining discriminative components with random forests. In: European conference on computer vision. pp. 446--461. Springer (2014)

\bibitem{cao2020heteroskedastic}
Cao, K., Chen, Y., Lu, J., Arechiga, N., Gaidon, A., Ma, T.: Heteroskedastic and imbalanced deep learning with adaptive regularization. arXiv preprint arXiv:2006.15766  (2020)

\bibitem{cao2019ldam}
Cao, K., Wei, C., Gaidon, A., Arechiga, N., Ma, T.: Learning imbalanced datasets with label-distribution-aware margin loss. NeurIPS  (2019)

\bibitem{chen2019understanding}
Chen, P., Liao, B.B., Chen, G., Zhang, S.: Understanding and utilizing deep neural networks trained with noisy labels. In: International Conference on Machine Learning. pp. 1062--1070. PMLR (2019)

\bibitem{chen2015webly}
Chen, X., Gupta, A.: Webly supervised learning of convolutional networks. In: Proceedings of the IEEE International Conference on Computer Vision. pp. 1431--1439 (2015)

\bibitem{chen2021boosting}
Chen, Y., Shen, X., Hu, S.X., Suykens, J.A.: Boosting co-teaching with compression regularization for label noise. In: Proceedings of the IEEE/CVF Conference on Computer Vision and Pattern Recognition. pp. 2688--2692 (2021)

\bibitem{cubuk2020randaugment}
Cubuk, E.D., Zoph, B., Shlens, J., Le, Q.V.: Randaugment: Practical automated data augmentation with a reduced search space. In: Proceedings of the IEEE/CVF Conference on Computer Vision and Pattern Recognition Workshops. pp. 702--703 (2020)

\bibitem{fawzi2016robustness}
Fawzi, A., Moosavi-Dezfooli, S.M., Frossard, P.: Robustness of classifiers: from adversarial to random noise. arXiv preprint arXiv:1608.08967  (2016)

\bibitem{frenay2013classification}
Fr{\'e}nay, B., Verleysen, M.: Classification in the presence of label noise: a survey. IEEE transactions on neural networks and learning systems  \textbf{25}(5),  845--869 (2013)

\bibitem{goodfellow2014generative}
Goodfellow, I., Pouget-Abadie, J., Mirza, M., Xu, B., Warde-Farley, D., Ozair, S., Courville, A., Bengio, Y.: Generative adversarial nets. NeurIPS  (2014)

\bibitem{han2018co}
Han, B., Yao, Q., Yu, X., Niu, G., Xu, M., Hu, W., Tsang, I., Sugiyama, M.: Co-teaching: Robust training of deep neural networks with extremely noisy labels. arXiv preprint arXiv:1804.06872  (2018)

\bibitem{he2016deep}
He, K., Zhang, X., Ren, S., Sun, J.: Deep residual learning for image recognition. In: Proceedings of the IEEE conference on computer vision and pattern recognition. pp. 770--778 (2016)

\bibitem{jamal2020rethinking}
Jamal, M.A., Brown, M., Yang, M.H., Wang, L., Gong, B.: Rethinking class-balanced methods for long-tailed visual recognition from a domain adaptation perspective. In: Proceedings of the IEEE/CVF Conference on Computer Vision and Pattern Recognition. pp. 7610--7619 (2020)

\bibitem{jiang2020beyond}
Jiang, L., Huang, D., Liu, M., Yang, W.: Beyond synthetic noise: Deep learning on controlled noisy labels. In: International Conference on Machine Learning. pp. 4804--4815. PMLR (2020)

\bibitem{jiang2018mentornet}
Jiang, L., Zhou, Z., Leung, T., Li, L.J., Fei-Fei, L.: Mentornet: Learning data-driven curriculum for very deep neural networks on corrupted labels. In: International Conference on Machine Learning. pp. 2304--2313. PMLR (2018)

\bibitem{kang2019decoupling}
Kang, B., Xie, S., Rohrbach, M., Yan, Z., Gordo, A., Feng, J., Kalantidis, Y.: Decoupling representation and classifier for long-tailed recognition. arXiv preprint arXiv:1910.09217  (2019)

\bibitem{karthik2021learning}
Karthik, S., Revaud, J., Chidlovskii, B.: Learning from long-tailed data with noisy labels. arXiv preprint arXiv:2108.11096  (2021)

\bibitem{kumar2020robust}
Kumar, H., Manwani, N., Sastry, P.: Robust learning of multi-label classifiers under label noise. Proceedings of the 7th ACM IKDD CoDS and 25th COMAD pp. 90--97 (2020)

\bibitem{lee2021learning}
Lee, J., Kim, E., Lee, J., Lee, J., Choo, J.: Learning debiased representation via disentangled feature augmentation. arXiv preprint arXiv:2107.01372  (2021)

\bibitem{lee2018cleannet}
Lee, K.H., He, X., Zhang, L., Yang, L.: Cleannet: Transfer learning for scalable image classifier training with label noise. In: Proceedings of the IEEE Conference on Computer Vision and Pattern Recognition. pp. 5447--5456 (2018)

\bibitem{li2020dividemix}
Li, J., Socher, R., Hoi, S.C.: Dividemix: Learning with noisy labels as semi-supervised learning. arXiv preprint arXiv:2002.07394  (2020)

\bibitem{li2020gradient}
Li, M., Soltanolkotabi, M., Oymak, S.: Gradient descent with early stopping is provably robust to label noise for overparameterized neural networks. In: International conference on artificial intelligence and statistics. pp. 4313--4324. PMLR (2020)

\bibitem{liu2020deep}
Liu, J., Sun, Y., Han, C., Dou, Z., Li, W.: Deep representation learning on long-tailed data: A learnable embedding augmentation perspective. In: CVPR (2020)

\bibitem{liu2020peer}
Liu, Y., Guo, H.: Peer loss functions: Learning from noisy labels without knowing noise rates. In: International Conference on Machine Learning. pp. 6226--6236. PMLR (2020)

\bibitem{liu2019large}
Liu, Z., Miao, Z., Zhan, X., Wang, J., Gong, B., Yu, S.X.: Large-scale long-tailed recognition in an open world. In: Proceedings of the IEEE/CVF Conference on Computer Vision and Pattern Recognition. pp. 2537--2546 (2019)

\bibitem{ma2020normalized}
Ma, X., Huang, H., Wang, Y., Romano, S., Erfani, S., Bailey, J.: Normalized loss functions for deep learning with noisy labels. In: International Conference on Machine Learning. pp. 6543--6553. PMLR (2020)

\bibitem{van2008visualizing}
Van~der Maaten, L., Hinton, G.: Visualizing data using t-sne. Journal of machine learning research  \textbf{9}(11) (2008)

\bibitem{menon2020long}
Menon, A.K., Jayasumana, S., Rawat, A.S., Jain, H., Veit, A., Kumar, S.: Long-tail learning via logit adjustment. arXiv preprint arXiv:2007.07314  (2020)

\bibitem{mirzasoleiman2020coresets}
Mirzasoleiman, B., Cao, K., Leskovec, J.: Coresets for robust training of neural networks against noisy labels. arXiv preprint arXiv:2011.07451  (2020)

\bibitem{northcutt2021confident}
Northcutt, C., Jiang, L., Chuang, I.: Confident learning: Estimating uncertainty in dataset labels. Journal of Artificial Intelligence Research  \textbf{70},  1373--1411 (2021)

\bibitem{qi2022class}
Qi, J., Tang, K., Sun, Q., Hua, X.S., Zhang, H.: Class is invariant to context and vice versa: On learning invariance for out-of-distribution generalization. In: ECCV (2022)

\bibitem{ren2020balanced}
Ren, J., Yu, C., Sheng, S., Ma, X., Zhao, H., Yi, S., Li, H.: Balanced meta-softmax for long-tailed visual recognition. arXiv preprint arXiv:2007.10740  (2020)

\bibitem{rolnick2017deep}
Rolnick, D., Veit, A., Belongie, S., Shavit, N.: Deep learning is robust to massive label noise. arXiv preprint arXiv:1705.10694  (2017)

\bibitem{rosenfeld2020risks}
Rosenfeld, E., Ravikumar, P., Risteski, A.: The risks of invariant risk minimization. arXiv preprint arXiv:2010.05761  (2020)

\bibitem{sastry2017robust}
Sastry, P., Manwani, N.: Robust learning of classifiers in the presence of label noise. In: Pattern Recognition and Big Data, pp. 167--197. World Scientific (2017)

\bibitem{shore1981properties}
Shore, J., Johnson, R.: Properties of cross-entropy minimization. IEEE Transactions on Information Theory  \textbf{27}(4),  472--482 (1981)

\bibitem{shu2019meta}
Shu, J., Xie, Q., Yi, L., Zhao, Q., Zhou, S., Xu, Z., Meng, D.: Meta-weight-net: Learning an explicit mapping for sample weighting. arXiv preprint arXiv:1902.07379  (2019)

\bibitem{song2019selfie}
Song, H., Kim, M., Lee, J.G.: {SELFIE}: Refurbishing unclean samples for robust deep learning. In: ICML (2019)

\bibitem{sun2019meta}
Sun, Q., Liu, Y., Chua, T.S., Schiele, B.: Meta-transfer learning for few-shot learning. In: Proceedings of the IEEE/CVF Conference on Computer Vision and Pattern Recognition. pp. 403--412 (2019)

\bibitem{tan2021co}
Tan, C., Xia, J., Wu, L., Li, S.Z.: Co-learning: Learning from noisy labels with self-supervision. In: Proceedings of the 29th ACM International Conference on Multimedia. pp. 1405--1413 (2021)

\bibitem{tang2020long}
Tang, K., Huang, J., Zhang, H.: Long-tailed classification by keeping the good and removing the bad momentum causal effect. arXiv preprint arXiv:2009.12991  (2020)

\bibitem{wang2022equivariance}
Wang, T., Sun, Q., Pranata, S., Jayashree, K., Zhang, H.: Equivariance and invariance inductive bias for learning from insufficient data. In: European Conference on Computer Vision (ECCV) (2022)

\bibitem{wang2021self}
Wang, T., Yue, Z., Huang, J., Sun, Q., Zhang, H.: Self-supervised learning disentangled group representation as feature. arXiv preprint arXiv:2110.15255  (2021)

\bibitem{wang2020long}
Wang, X., Lian, L., Miao, Z., Liu, Z., Yu, S.X.: Long-tailed recognition by routing diverse distribution-aware experts. ICLR  (2020)

\bibitem{wei2020combating}
Wei, H., Feng, L., Chen, X., An, B.: Combating noisy labels by agreement: A joint training method with co-regularization. In: Proceedings of the IEEE/CVF Conference on Computer Vision and Pattern Recognition. pp. 13726--13735 (2020)

\bibitem{wei2021open}
Wei, H., Tao, L., Xie, R., An, B.: Open-set label noise can improve robustness against inherent label noise. arXiv preprint arXiv:2106.10891  (2021)

\bibitem{wei2021robust}
Wei, T., Shi, J.X., Tu, W.W., Li, Y.F.: Robust long-tailed learning under label noise. arXiv preprint arXiv:2108.11569  (2021)

\bibitem{wu2020topological}
Wu, P., Zheng, S., Goswami, M., Metaxas, D., Chen, C.: A topological filter for learning with label noise. arXiv preprint arXiv:2012.04835  (2020)

\bibitem{yu2019does}
Yu, X., Han, B., Yao, J., Niu, G., Tsang, I., Sugiyama, M.: How does disagreement help generalization against label corruption? In: International Conference on Machine Learning. pp. 7164--7173. PMLR (2019)

\bibitem{zhang2017mixup}
Zhang, H., Cisse, M., Dauphin, Y.N., Lopez-Paz, D.: mixup: Beyond empirical risk minimization. arXiv preprint arXiv:1710.09412  (2017)

\bibitem{zhang2021test}
Zhang, Y., Hooi, B., Hong, L., Feng, J.: Test-agnostic long-tailed recognition by test-time aggregating diverse experts with self-supervision. arXiv preprint arXiv:2107.09249  (2021)

\bibitem{zhang2021deep}
Zhang, Y., Kang, B., Hooi, B., Yan, S., Feng, J.: Deep long-tailed learning: A survey. arXiv preprint arXiv:2110.04596  (2021)

\bibitem{zhou2020bbn}
Zhou, B., Cui, Q., Wei, X.S., Chen, Z.M.: Bbn: Bilateral-branch network with cumulative learning for long-tailed visual recognition. In: Proceedings of the IEEE/CVF Conference on Computer Vision and Pattern Recognition. pp. 9719--9728 (2020)

\end{thebibliography}
\end{document}